\documentclass{article}
\usepackage{latexsym}
\usepackage{amsmath}
\usepackage{amssymb}
\usepackage{a4wide}
\usepackage{amsthm}
\usepackage{url}
\usepackage{lscape}
\usepackage[table]{xcolor}
\usepackage{graphicx}
\usepackage{dirtytalk}
\usepackage{subcaption}
\usepackage{url}
\usepackage{natbib}
\usepackage{pdflscape}

\DeclareMathOperator*{\atan}{atan}
\DeclareMathOperator*{\hypot}{hypot}
\DeclareMathOperator*{\NB}{NB}
\DeclareMathOperator*{\MN}{MN}
\DeclareMathOperator*{\NC}{NC}

\def\supertiny{ \font\supertinyfont = cmr10 at 8pt \relax
\supertinyfont}

\begin{document}

\title{Selection Heuristics on Semantic Genetic Programming for Classification Problems}  
\author{Claudia N. S\'anchez$^{1,2}$ \and Mario~Graff$^{1,3}$}%

\date{$^1$INFOTEC Centro de Investigaci\'on e Innovaci\'on en Tecnolog\'ias
de la Informaci\'on y Comunicaci\'on, Circuito Tecnopolo Sur No 112, Fracc. Tecnopolo Pocitos II, Aguascalientes 20313, M\'exico\\
$^2$Facultad de Ingenier\'ia. Universidad Panamericana, Aguascalientes, M\'exico\\
$^3$CONACyT Consejo Nacional de Ciencia y Tecnolog\'ia,
Direcci\'on de Cátedras, Insurgentes Sur 1582, Cr\'edito Constructor, Ciudad de M\'exico 03940 M\'exico~\\ ~\\
This work has been submitted to the Evolutionary Computation journal for possible publication.}

\maketitle

\begin{abstract}

Individual's semantics have been used for guiding the learning process of Genetic Programming solving supervised learning problems. The semantics has been used to proposed novel genetic operators as well as different ways of performing parent selection. The latter is the focus of this contribution by proposing three heuristics for parent selection that replace the fitness function on the selection mechanism entirely. These heuristics complement previous work by being inspired in the characteristics of the addition, Naive Bayes, and Nearest Centroid functions and applying them only when the function is used to create an offspring. These heuristics use different similarity measures among the parents to decide which of them is more appropriate given a function. The similarity functions considered are the cosine similarity, Pearson's correlation, and agreement. We analyze these heuristics' performance against random selection, state-of-the-art selection schemes, and 18 classifiers, including auto-machine-learning techniques, on 30 classification problems with a variable number of samples, variables, and classes. The result indicated that the combination of parent selection based on agreement and random selection to replace an individual in the population produces statistically better results than the classical selection and state-of-the-art schemes, and it is competitive with state-of-the-art classifiers. Finally, the code is released as open-source software. 
\end{abstract}

\section{Introduction}

Classification is a supervised learning problem that consists of finding a function that learns the relation between inputs and outputs, where the outputs are a set of labels. It could be applied for solving problems like object recognition, medical diagnosis, identification of symbols, among others. The starting point would be the training set composed of input-output pairs, i.e., $\mathcal X = \{ (\vec{x_1}, y_1), \ldots, (\vec{x_n}, y_n)\}$, where $\forall_{\vec{x} \in \mathcal{X}} \; \vec{x} \in \mathbb{R}^m$, and $\forall_{y \in \mathcal{X} } \; y \in \mathbb{R}$. The training set $\mathcal{X}$ is used to find a function, $f$, that minimize a loss function, $\mathbb{L}$, that is, $f$ is the function that minimize $\sum_{(\vec{x}, y) \in \mathcal X} \mathbb{L}(f(\vec{x}), y)$, where the ideal scenario would be $\forall_{(\vec{x}, y) \in \mathcal X} f(\vec{x}) = y$, and, also to accurately predict the labels of unseen inputs. Genetic programming (GP) is a flexible and powerful evolutionary technique with some features that can be very valuable and suitable for the evolution of classifiers \citep{Espejo2010AClassification}. The first document related to GP was presented by \cite{friedberg1958learning}, but the term was coined by \cite{Koza1992GeneticSelection}. The classifiers constructed with GP started appearing around the 2000s \citep{Loveard2001RepresentingProgramming, brameier2001comparison}, and nowadays, they can be comparable with state-of-the-art machine learning techniques for solving hard classification problems. For example, GP has been used for medical proposes \citep{brameier2001comparison, Olson2016AutomatingOptimization}, image classification \citep{iqbal2017cross}, fault classification \citep{guo2005feature}, only to mention some of them. Also, the GP performance has been successfully compared with others machine learning techniques as Neural Networks \citep{brameier2001comparison}, Support Vector Machines \citep{lichodzijewski2008managing, mcintyre2011classification}. Specificaly, GP can be applied in the preprocessing task \citep{guo2005feature, badran2012multi, Ingalalli2014AProblems, LaCava2019MultidimensionalClassification}, in model extraction \citep{Loveard2001RepresentingProgramming, brameier2001comparison, Muni2004AProgramming, zhang2006using, folino2008training, mcintyre2011classification, Graff2016EvoDAG:Library, iqbal2017cross}, or for building machine learning pipelines \citep{Olson2016AutomatingOptimization}.

Traditionally, GP uses individuals' fitness to select the parents that build the next generation of individuals \citep{poli08:fieldguide}. Recently, new approaches for parent selection that use angles among individuals' semantics have been developed. For example, Angle-Driven Selection (ADS), proposed by \cite{Chen2019ImprovingOperators}, chooses parents maximizing the angle between their relative semantics aiming to have parents with different behaviors. \cite{vanneschi2019alignment} introduced a selection scheme based on the angle between the error vectors. Besides, there have been approaches that analyze evolutive algorithms' behavior where the fitness function is replaced with a heuristic. \cite{Nguyen2012AnMetrics} proposed Fitness Sharing, a technique that promotes dispersion and diversity of individuals. \cite{Lehman2011AbandoningAlone} proposed Novelty Search, where the individual's fitness is related totally to its novelty without care about its target behavior. The individual's novelty is computed as the average distance between its behavior and its $k$-nearest neighbors' behavior. \cite{Naredo2016EvolvingSearch} used Novelty Search in GP for solving classification problems. However, one of the main characteristics of GP is that the function set defines its search space, and to the best of our knowledge, there are not documents that propose the use of functions' properties for parent selection.

The intersection between our proposal and the previous research works is on the selection stage of the evolutionary process. Most of the proposed methods produce a selection mechanism that considers the semantics of individuals as well as the individual's fitness to choose the parents. The exception is the proposals using Novelty Search, where the selection process is entirely performed using the semantics. It enhances population diversity. Our approach follows this path by abandoning the fitness function in the selection process. The difference with Novelty Search is that our proposal searches for individuals whose semantics might help the function used to create the offspring. Besides, once the individual was created, local optimization is performed using the training outputs, or, traditionally named, the target semantics. Functions' properties inspire our selection heuristics; in particular, these were inspired by the addition and the classifiers Naive Bayes and Nearest Centroid. The proposed heuristics measure the similarity among parents. They are based on cosine similarity, Pearson's correlation coefficient, and agreement\footnote{We thank the anonymous reviewer for suggesting this name which considerably improves the description clarity.}. Specifically, our heuristic based on Pearson's correlation coefficient is quite similar to ADS, but the difference is that our proposal is applied only to specific functions. In this document, we present the comparison of the use of ADS and our proposal. 

The selection heuristics are tested on a steady-state GP called EvoDAG. It was inspired by the Geometric Semantic genetic operators proposed by \cite{Moraglio2012GeometricProgramming} with the implementation of Vanneschi {\it et al.} \citep{vanneschi2013new,Castelli2015AProgramming}. It has been successfully applied to a variety of text classification problems \citep{Graff2018EvoMSA:Analysis}. In steady-state evolution, the selection mechanism is used twice to perform parent selection and decide the individual being replaced by an offspring; we refer to the later negative selection. The proposed selection is used to select the parent; however, on the negative selection, we also tested the system's performance using random selection or the traditional selection guided by fitness. 

We analyze the performance of different GP systems obtained by combining the schemes for parent selection (our heuristics, random selection, and traditional selection) and negative selection (random and traditional selection). Besides, we compare our selection heuristics against two state-of-the-art techniques, Angle-Driven-Selection \citep{Chen2019ImprovingOperators}, and Novelty Search \citep{Naredo2016EvolvingSearch}. To provide a complete picture of the performance of our selection heuristics, we decided to compare them against state-of-the-art classifiers and two auto-machine learning algorithms. The results show that our selection heuristic guided by agreement and random negative selection outperforms traditional selection, presenting a statistically significant difference. On the other hand, our selection heuristics outperforms the state-of-the-art techniques Angle-Driven-Selection and Novelty Search. Furthermore, compared with state-of-the-art classifiers, our GP system obtained the second-lowest rank being the first one TPOT, an auto-machine learning technique, although the difference in performance between these systems, in terms of macro-F1, is not statistically significant.

The rest of the manuscript is organized as follows. Section \ref{sec:related_work} presents the related work. The GP system used to test the proposed heuristics is described in Section \ref{sec:evodag}. The proposed selection heuristics are presented in Section \ref{sec:selection}. Section \ref{sec:results} presents the experiments and results. The discussion and limitations of the approach are treated in Section \ref{sec:discussion}. Finally, Section \ref{sec:conclusions} concludes this research.

\section{Related work}
\label{sec:related_work}

We can define supervised learning as follows \cite{Vanneschi2017AnProgramming}: given the training set composed of input-output pairs, $\mathcal X = \{ (\vec{x_1}, y_1), \ldots, (\vec{x_n}, y_n)\}$, where $\forall_{\vec{x} \in \mathcal{X}} \; \vec{x} \in \mathbb{R}^m$, and $\forall_{y \in \mathcal{X} } \; y \in \mathbb{R}$; the learning process can be defined as the problem of finding a function $f$ that minimizes a loss function, $\mathbb{L}$, that is, $f$ is the function that minimize $\sum_{(\vec{x}, y) \in \mathcal X} \mathbb{L}(f(\vec{x}), y)$, where the ideal scenario would be $\forall_{(\vec{x}, y) \in \mathcal X} f(\vec{x}) = y$. Traditionally, the vector composed by all the outputs, $\vec{t} = \{ y_1, \ldots, y_n \}$, is called target vector. In this way, a GP individual $P$ can be seen as a function $h$ that, for each input vector $\vec{x_i}$ returns the scalar value $h(\vec{x_i})$, and the objective is to find the GP individual that minimizes $\sum_{ (\vec{x}, y) \in \mathcal X} \mathbb{L}(h(\vec{x}), y)$.

On Semantic Genetic Programming (SGP), for example in \citep{Vanneschi2017AnProgramming}, each individual $P$ can be represented by its semantics vector $\vec{S_p}$ that corresponds to all the inputs' evaluations in the function $h$, this is, $\vec{S_p} = \{ h(\vec{x_1}), \ldots, h(\vec{x_n})\}$. We can imagine the existence of two spaces: the genotype space, where individuals are represented by their structures, and the phenotype or semantic space, where individuals are represented by points, which are their semantics. Remark that the target vector $\vec{t}$ is a point in the semantic space. The dimensionality of the semantic space is the number of input vectors. Using this notation, the objective of SGP is to find the individual $P$ whose semantics $S_p$ is as close as possible to the target vector $\vec{t}$ in the semantic space. The distance between the individual's semantics $\vec{S_p}$ and the target vector $\vec{t}$ is used as the individual's fitness.

In order to provide a complete picture of the research documents related to this contribution, the section starts by describing semantic genetic operators; this is followed by presenting research proposals proposing selection mechanisms; and, lastly, some proposals where GP has been used to develop classifiers. 

\subsection{Semantic Genetic Operators}

Semantic Genetic Programming uses the target behavior, $\vec{t}$, to guide the search. \cite{Krawiec2016SemanticProgramming} affirmed that semantically aware methods make search algorithms better informed. Specifically, several crossover and mutation operators have been developed with the use of semantics. \cite{Beadle2008SemanticallyProgramming} proposed a crossover operator that measures the semantic equivalence between parents and offsprings and rejects the offspring that is semantically equivalent to its parents. \cite{Uy2011Semantically-basedRegression} developed a semantic crossover and a mutation operator. The crossover operator searches for a crossover point in each parent so that subtrees are semantically similar, and the mutation operator allows the replacement of an individual's subtree only if the new subtree is semantically similar. \cite{Hara2012NewProgramming} proposed the Semantic Control Crossover that uses semantics to combine individuals. A global search is performed in the first generations and a local search in the last ones. Graff {\it et al.} used subtrees semantics and partial derivatives to propose a crossover \citep{Graff2014SemanticError, Suarez2015SemanticFunction} and a mutation \citep{Graff2014GeneticError} operator.

Moraglio {\it et al.} proposed the Geometric Semantic Genetic Programming (GSGP) \citep{Moraglio2004TopologicalCrossover, Moraglio2012GeometricProgramming}. Their work called the GP scientific community's attention because the crossover operator produces an offspring that stands in the segment joining the parents' semantics. Therefore, offspring fitness cannot be worse than the worst fitness of the parents. Given two parents' semantics $\vec{p_1}$ and $\vec{p_2}$, the crossover operator generates an offspring whose semantics is $r \cdot \vec{p_1} + (1-r) \cdot \vec{p_2}$, where $r$ is a real value between $0$ and $1$. This property transforms the fitness landscape into a cone. Unfortunately, the offspring is always bigger than the sum of its parents' size; this makes the operator unusable in practice. Later, some operators were developed to improve Moraglio's GSGP. For example, Approximately Geometric Semantic Crossover (SX) \citep{Krawiec2009ApproximatingSpace}, Deterministic Geometric Semantic Crossover \citep{Hara2012NewProgramming}, Locally Geometric Crossover (LGX) \citep{Krawiec2012LocallyCrossover,Krawiec2013LocallyOperators},  Approximated Geometric Crossover (AGX) \citep{Pawlak2015SemanticProgramming}, and Subtree Semantic Geometric Crossover (SSGX) \citep{Nguyen2016SubtreeProgramming}. 

\cite{Graff2015SemanticSpace} proposed a new crossover operator based on projections in the phenotype space. It creates a plane in the semantic space using the parents' semantics. The offspring is calculated as the projection of the target in that plane. Given the parents' semantics $\vec{p_1}$ and $\vec{p_2}$, and the target semantics $\vec{t}$, the offspring is calculated as $\alpha \vec{p_1} + \beta \vec{p_2}$, where $\alpha$ and $\beta$ are real values that are calculated solving the equation $A [\alpha,\beta]'=\vec{t}$, where $A=(\vec{p_1},\vec{p_1})$. It implies the offspring will be at least as good as the best parent. Memetic Genetic Programming based on Orthogonal Projections in the Phenotype Space was also proposed by \cite{Graff2015MemeticSpace}. In that work, they used a linear combination of $k$ parents as $\sum_k \alpha_k \vec{p_k}$, where $\vec{p_k}$ represents the semantics of the $k$ parent. The main idea is optimizing the coefficients $\{\alpha_i\}$ with ordinary least squares (OLS) to guarantee that the offspring is the best of its family. As a result, the generated tree's fitness is always better or equal to any internal tree. It was not the first time that parameters were added to GP nodes; \cite{smart2004continuously} defined the Inclusion Factors as numeric values between 0 and 1 assigned to each node in the tree structure, except the root node. This value represents the inclusion proportion of the node in the tree. \cite{Castelli2015GeometricSearch} presented a mutation operator, in their work called Geometric Semantic Genetic Programming with Local Search, based on Moraglio's mutation operator, that also uses parameters. In this operator, an individual's semantics $\vec{p}$ is modified with the following equation, $\alpha_0 + \alpha_1 \vec{p} + \alpha_2 (\vec{r_1}-\vec{r_2})$, where $\vec{r_1}$ and $\vec{r_2}$ are the semantics of random trees, and, $\alpha_i \; \epsilon \; \mathbb{R}$. $\{\alpha_i\}$ are calculated using the target semantics and OLS for getting the better linear combination of the individual's semantics and the random trees. They extended their work in \citep{castelli2019extending} applying Local Search to all the individuals during a separate step after mutation and crossover. For each individual, $p$, they calculated another one as $p' = \alpha p + \beta$, where $\alpha$ and $\beta$ are optimized with OLS minimizing the error between the individual's semantics and target semantics. Moreover, they generalized the idea and transformed it into a regression problem $p' = \sum_j \alpha_j f_j (p)$, where $f_j:\mathbb{R} \rightarrow \mathbb{R}$.

All those operators use semantics to guide the learning process; however, one of the main characteristics of GP is that the function set defines its search space, and to the best of our knowledge, there are not documents that propose the use of functions' properties for designing operators.

\subsection{Fitness and Selection in Genetic Programming}
\label{subsection:relatedwork_fitness_selection}

According to \cite{Vanneschi2014AProgramming}, one way to promote diversity in GP is by using different selection schemes. \cite{Nguyen2012AnMetrics} proposed Fitness Sharing, a technique that promotes dispersion and diversity of individuals. Their proposal consisted of calculating an individual's shared fitness as $f'_i=f_i (m_i+1)$, where $f_i$ is the individual's fitness, and $m_i$ is approximately equal to the number of individuals that behave similarly to individual $i$. \cite{Galvan-Lopez2013UsingDiversity} applied crossover only to those individuals whose difference in behavior is greater than a defined threshold for every element on the semantic vectors. Hara {\em et al.} proposed Deterministic Geometric Semantic Crossover \citep{Hara2012NewProgramming}, and later, they proposed to select the parents in such a way that the line connecting them is as close as possible to the target in the semantic space \citep{Hara2016DeterministicSelection}.

\cite{Ruberto2014ESAGPSpace} defined the Error Vector and Error Space. The individual error vector $\vec{e_p}$ is defined as $\vec{e_p}= \vec{p} - \vec{t}$, where $\vec{p}$ is the individual's semantics and $\vec{t}$ represents the target semantics. The error space contains all the individuals represented by their error vectors, where $\vec{t}$ is the origin. The proposal is to search, in the error space, for two or three individuals aligned, instead of using the fitness function; the rationality comes from the fact that given the aligned individuals, then there is a straightforward procedure to compute the optimal solution. \cite{Chu2016TournamentProgramming,Chu2018SemanticVectors} used the error vectors and the Wilcoxon signed-rank test to decide whether to select the fittest or the smaller individual as a parent. Their results show that the proposed techniques aim to enhance semantic diversity and reduce the code bloat in GP. \cite{vanneschi2019alignment} introduced a selection scheme based on the angle between the error vectors.

\cite{Chen2019ImprovingOperators} proposed Angle-Driven Geometric Semantic Genetic Programming (ADGSGP). Their work attempts to further explore the geometry of geometric operators in the search space to improve GP for symbolic regression. Their proposal included Angle-Driven Selection (ADS) that selects a pair of parents that have good fitness values and are far away from each other regarding the angle-distance of their relative semantics. The first parent is selected using fitness, and the second one is chosen maximizing the relative angle-distance between its semantics and the semantics of the first parent. The angle-distance is defined as $\gamma_r =  \arccos ( \frac{(\vec{t}-\vec{p_1})(\vec{t}-\vec{p_2})}{ \mid \mid \vec{t} - \vec{p_1} \mid \mid \mid \mid \vec{t}-\vec{p_2} \mid \mid})$, where $\vec{t}$, $\vec{p_1}$, and $\vec{p_2}$, represent the target semantics, and semantics of the first and the second parent, respectively. Also, they proposed Perpendicular Crossover (PC) and Random Segment Mutation (RSM) that likewise used angles to guide their process. Their experiments show that the angle-driven geometric operators drive the evolutionary process to fit the target semantics more efficiently and improve the generalization performance.

Our proposal, as well as these documents, aims to promote individuals' diversity but uses functions' properties for guiding the parent selection.

\subsection{Genetic Programming for classification}

Genetic programming (GP) is a flexible and powerful evolutionary technique with some features that can be very valuable and suitable for the evolution of classifiers \citep{Espejo2010AClassification}. For example, GP has been used for medical proposes \citep{brameier2001comparison, Olson2016AutomatingOptimization}, image classification \citep{iqbal2017cross}, and fault classification \citep{guo2005feature}. Also, GP classifiers have been successfully compared with state-of-the-art classifiers. \cite{brameier2001comparison} compared GP against neural networks on medical classification problems from a benchmark database. Their results show that GP performs comparably in classification and generalization. \cite{mcintyre2011classification} compared their GP framework against SVM classifiers over 12 UCI datasets with between 150 and 200,000 training instances. Solutions from the GP framework appear to provide a good balance between classification performance and model complexity, especially as the dataset instance count increases. \cite{LaCava2019MultidimensionalClassification} compared their GP framework to several state-of-the-art classification techniques (Random Forest, Neural Networks, and Support Vector Machines) across a broad set of problems, and showed that their technique achieves competitive test accuracies while also producing concise models.

Data can be transformed at the preprocessing stage to increase the quality of the knowledge obtained, and GP can be used to perform this transformation \citep{Espejo2010AClassification}. \cite{badran2012multi} proposed multi-objective GP to evolve a feature extraction stage for multiple-class classifiers. They found mappings that transform the input space into a new multi-dimensional decision space to increase the discrimination between all classes; the number of dimensions of this decision space is optimized as part of the evolutionary process. \cite{Ingalalli2014AProblems} introduced a GP framework called Multi-dimensional Multi-class Genetic Programming (M2GP). The main idea is to transform the original space into another one using functions evolved with GP, then, a centroid is calculated for each class, and the vectors are assigned to the class that corresponds to the nearest centroid using the Mahalanobis distance. M2GP takes as an argument the dimension of the transformed space. This parameter is evolved in M3GP \citep{Munoz2015M3GPGP} by including specialized search operators that can increase or decrease the number of feature dimensions produced by each tree. They extended M3GP and proposed M4GP \citep{LaCava2019MultidimensionalClassification} that uses a stack-based representation in addition to new selection methods, namely lexicase selection and age-fitness Pareto survival. 

\cite{Naredo2016EvolvingSearch} used Novelty Search (NS) for evolving GP classifiers based on M3GP, where the difference is the procedure to compute the fitness. Each GP individual is represented as a binary vector whose length is the training set size, and each vector element is set to 1 if the classifier assigns the class label correctly and 0 otherwise. Then, those binary vectors are used to measure the sparseness among individuals, and the more the sparseness, the higher the fitness value. Their results show that all their NS variants achieve competitive results relative to the traditional objective-based.

Auto machine learning consists of obtaining a classifier (or a regressor) automatically. It includes the steps of preprocessing, feature selection, classifier selection, and hyperparameters tuning. \cite{Feurer2015EfficientLearning} developed a robust automated machine learning (AutoML) technique using Bayesian optimization methods. It is based on scikit-learn \citep{Pedregosa2011Scikit-learn:Python}, using 15 classifiers, 14 feature preprocessing methods, and 4 data preprocessing methods; giving rise to a structured hypothesis space with 110 hyperparameters. \cite{Olson2016AutomatingOptimization} proposed the use of GP to develop an algorithm that automatically constructs and optimizes machine learning pipelines through a Tree-based Pipeline Optimization Tool (TPOT). On classification, the objective consists of maximizing the accuracy score performing a searching of the combinations of 14 preprocessors, five feature selectors, and 11 classifiers; all these techniques are implemented on scikit-learn. It is interesting to note that TPOT uses a tree-based GP programming approach where the different learning process components are nodes in a tree, and traditional subtree crossover is used as a genetic operator. 

The use of the proposed selection heuristics improves EvoDAG, a GP system described in Section \ref{sec:evodag}, making it competitive with the auto-machine learning techniques explained above and other state-of-the-art classifiers.

\section{Genetic Programming System}
\label{sec:evodag}

We decided to implement the proposed selection heuristics and the selection heuristics of the state-of-the-art in our previously developed GP system called EvoDAG\footnote{https://github.com/mgraffg/EvoDAG} \citep{Graff2016EvoDAG:Library, Graff2017SemanticAnalysis}. EvoDAG is inspired by the implementation of GSGP performed by \cite{Castelli2015AProgramming}, where the main idea is to keep track of all the individuals and their behavior, leading to an efficient evaluation of the offspring whose complexity depends only on the number of fitness cases. Let us recall that the offspring, in the geometric semantic crossover, is $\vec{o} = r \vec{p_1} + (1 - r) \vec{p_2}$, where $r$ is a random function or a constant, and, $\vec{p_1}$ and $\vec{p_2}$ are the parents' semantics. As it was explained in the previous section, in \citep{Graff2015SemanticSpace}, we decided to extend this operation by allowing the offspring to be a linear combination of the parents, that is, $\vec{o} = \theta_1 \vec{p_1} + \theta_2 \vec{p_2}$, where $\theta_1$ and $\theta_2$ are obtained using ordinary least squares (OLS) minimizing the difference between the offspring and the target semantics. Continuing with this line of research, in \citep{Graff2016EvoDAG:Library}, we investigated the case when the offspring is a linear combination of more than two parents, and, also, to include the possibility that the parents could be combined using a function randomly selected from the function set.

EvoDAG, as customary, uses a function set $\mathcal{F} = \{ \sum_{60}$, $\prod_{20}$, $\max_5$, $\min_5$, $\sqrt{\cdot}$, $\mid \cdot \mid$, $\sin$, $\tan$, $\atan$, $\tanh$, $\hypot_2$, $\NB_5$, $\MN_5$, $\NC_2\}$, and a terminal set $\mathcal T = \{x_1, \ldots, x_m \}$, to create the individuals. It is also included in $\mathcal F$ classifiers such as Naive Bayes with Gaussian distribution ($\NB_5$), with Multinomial distribution ($\MN_5$), and Nearest Centroid ($\NC_2$). The function-set elements are traditional operations where the subscript indicates the number of arguments. EvoDAG's default parameters, including the number of arguments, were defined performing a random search \citep{Bergstra2012RandomOptimization}, on the parameter space, using as a benchmark of classification problems that included different sentiment analysis problems as well as problems taken from UCI repository (not included in the problems used to measure the performance of the selection heuristics). The final values were the consensus of the parameters obtaining the best performance in the problems tested.

The initial population starts with $\mathcal P = \{\theta_1 x_1, \ldots, \theta_m x_m$, $\NB(x_1, \ldots, x_m)$, $\MN(x_1, \ldots, x_m)$, $\NC(x_1, \ldots, x_m)\}$, where $x_i$ is the $i$-th input, and $\theta_i$ is obtained using OLS. In the case that the number of individuals is lower than the population size, the process starts including an individual created by randomly selecting a function from $\mathcal F$ and the arguments are drawn from the current population $\mathcal P$. For example, let $\hypot$ be the selected function, and the first and second arguments are $\theta_2 x_2$, and $\NB(x_1, \ldots, x_m)$. Then, the individual inserted to $\mathcal P$ is $\theta \hypot(\theta_2 x_2, \NB(x_1, \ldots, x_m))$, where $\theta$ is obtained using OLS. This process continues until the population size is reached; EvoDAG sets population size of $4000$.

EvoDAG uses a steady-state evolution; consequently, $\mathcal P$ is updated by replacing a current individual, selected using a negative selection, with an offspring that can be selected as a parent just after being inserted in $\mathcal P$. The evolution process is similar to the one used to create the initial population, and the difference is in the procedure used to select the arguments. That is, a function $f$ is selected from $\mathcal F$, its arguments are selected from $\mathcal P$ using tournament selection or any of the proposed selection heuristics, and finally, the parameters $\theta$ associated to $f$ are optimized using OLS. The addition is defined as $ \sum_i \theta_i x_i$, where $x_i$ is an individual in $\mathcal P$. The rest of the arithmetic functions, trigonometric functions, $\min$, and $\max$ are defined as $\theta f(\ldots, x_i, \ldots)$, where $f$ is the function at hand, and $x_i$ is an individual in $\mathcal P$. For preventing overfitting, EvoDAG stops the evolutionary process using early stopping; that is, the training set is split into a smaller training set (50\% reduction) and a validation set containing the remaining elements. The training set is used to calculate the fitness and the parameters $\theta$. The evolution stops when the best individual on the validation set has not been updated in a defined number of evaluations; EvoDAG sets this as $4000$. The final model corresponds to the best individual in the validation set found during the whole evolutionary process.  

\begin{figure}[htb!]
  \centering
    \includegraphics[width=0.8\textwidth]{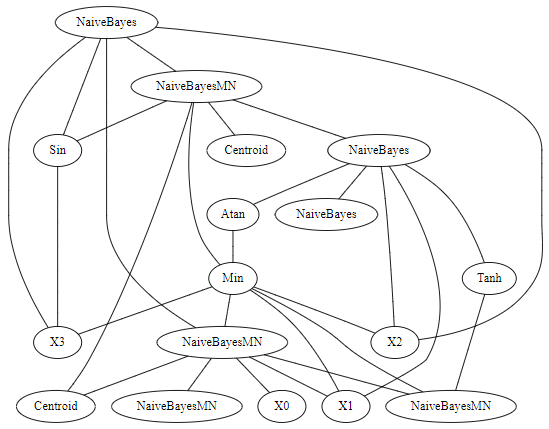}
  \caption{A model evolved by EvoDAG on the Iris dataset. The inputs are in the bottom of the figure and the output is on the top.}
  \label{fig:evodag_model}
\end{figure}

At this point, it is worth mentioning that EvoDAG uses a one-vs-rest scheme on classification problems. That is, a problem with $k$ different classes is converted into $k$ problems; each one assigns $1$ to the current class and $-1$ to the other labels. Instead of evolving one tree per class, as done, for example, in \cite{Muni2004AProgramming}, we decided to use only one tree and optimize $k$ different $\theta$ parameters, one for each label. The result is that each node outputs $k$ values, and the class is the one with the highest value. In the nodes representing classifiers, like Naive Bayes or Nearest Centroid, the output is the log-likelihood. In order to provide an idea of the type of models produced by EvoDAG, Figure \ref{fig:evodag_model} presents a model of the Iris dataset. The inputs ($x_0,\ldots, x_3, \NB, \MN, \NC$) are at the bottom of the figure. The computation flow goes from bottom to top; the output node is at the top of the figure, i.e., Naive Bayes using Gaussian distribution. The figure helps to understand the role of optimizing the $k$ set of parameters, one for each class, where each node outputs $k$ values; consequently, each node is a classifier. EvoDAG uses macro-F1 score to calculate the individuals' fitness. The macro-F1 score was chosen because it helps to handle imbalanced datasets. The class imbalance problem typically occurs when, in a classification problem, there are many more instances of some classes than others. In such cases, standard classifiers tend to be overwhelmed by the class with more examples ignoring the less represented classes \citep{Chawla2004Editorial:Sets}.  

It is well known that in evolutionary algorithms, there are runs that do not produce an acceptable result, to improve stability, we decided to use Bagging \citep{Breiman1996BaggingPredictors} in our approach. We create $30$ different models by randomly selecting $50\%$ samples for training and the remaining for validation. A bagging estimator can be expected to perform similarly by either drawing $n$ elements from the training set with-replacement or selecting $\frac{n}{2}$ elements without-replacement \citep{Friedman2007OnEstimation}. In addition, we reduce the learning complexity that, in EvoDAG's case, is measured in terms of training samples. EvoDAG's final prediction is the average of the models' predictions.

\section{Selection Heuristics}
\label{sec:selection}

This document proposes selection heuristics for GP tailored to classification problems based on the idea that functions' properties and individuals' semantics can guide parent selection. The heuristics replace the fitness function used in the selection procedure - tested in particular in tournament selection - to select the parent. Let us recall that in a steady-state evolution, there are two stages where selection takes place. On the one hand, the selection is used to choose the parents, and on the other hand, the selection is applied to decide which individual, in the current population, is replaced by the offspring. This last one is called negative selection. The most popular selection method in GP is tournament selection \citep{Fang2010AProgramming}, and also, negative selection is commonly performed using the same selection scheme; nonetheless, in the latter case, the winner of the tournament is the one with the worst fitness. In the rest of the section, we describe the process of creating an offspring in EvoDAG, and the traditional tournament selection (based on fitness), random selection, and our three proposed selection heuristics. 

\begin{figure}[htb!]
\centering
\includegraphics[width=1.0\textwidth]{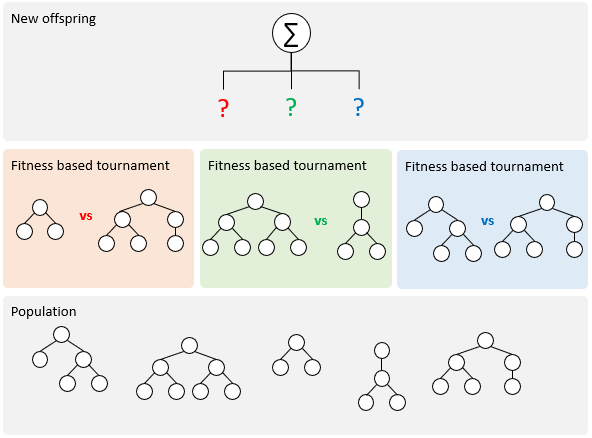}
 \caption[Diagram of parent selection based on fitness (in EvoDAG)]{ Diagram of tournament selection using the fitness function to decide the winner of the tournament.} 
  \label{fig:parent_selection_fit}
\end{figure}

In EvoDAG, the process of creating an offspring starts by selecting a function from the function set $\mathcal{F}$, and then parent selection needs to be performed to choose each one of the $k$ arguments (or parents). Figure \ref{fig:parent_selection_fit} shows an example of the traditional tournament selection (with tournament size two) where the function $\sum$ was selected, and $3$ individuals need to be selected as arguments from the population $\mathcal{P}$ for creating the new offspring. It can be seen that for selecting each argument, a binary tournament needs to be performed. For selecting an argument, two individuals from the population are randomly chosen, and the one with the highest fitness is selected as the argument; each tournament is depicted using a different color. The procedure represented in Figure \ref{fig:parent_selection_fit} also helps to describe {\bf random selection} where each argument of the function $\sum$ is selected randomly from the population, and a tournament is not needed. As can be seen, random selection is the most straightforward and less expensive strategy given that there is no need to perform the tournament.

Let us start by describing the selection heuristics that were inspired by functions' properties. The functions are the addition and the classifiers Naive Bayes and Nearest Centroid. The addition is defined in our GP system as $\sum_k \theta_k p_k$, where OLS and target semantics are used to estimate $ \theta_k$. As can be seen, to accurately identify the $k$ coefficients, the exogenous variables $p_k$ must be linearly independent. In general, knowing whether a set of vectors is linearly independent requires a non-zero determinant; however, the trivial case is when these vectors are orthogonal. Based on \cite{Brereton2016OrthogonalityVectors}, uncorrelated vectors are also linearly independent. In Naive Bayes's case, its model assumes that given a class, the features are independent. As expected, the process to calculate independence is expensive, so instead, we use the correlation among features in our heuristic. The correlation of two statistically independent variables is zero, although the inverse is not necessarily true. Finally, Nearest Centroid (NC) is a classifier representing each class with its centroid calculated with the elements associated with that class. The label of a given instance corresponds to the class of the closest centroid. Therefore, we think it might improve the performance of NC if the diversity of the inputs is increased. 

To sum up, the three functions (addition, Naive Bayes, and Nearest Centroid) perform better when their input vectors are orthogonal, uncorrelated, or independent. The main idea is quite similar to the proposal in Novelty Search \citep{Lehman2011AbandoningAlone}, where instead of promoting fitness, they promote diversity. In our case, we want fit individuals for the final solution, but we select parents promoting diversity among them. The proposed selection heuristics use the individuals' semantics for choosing diverse parents.  

The first selection heuristic ideally would select orthogonal vectors; evidently, this event is unlikely, so a function that measures the closeness to orthogonality is needed. The cosine similarity is such a measure, it is defined in Equation \ref{eq:cosine_similarity}, where $\vec{v_1}$ and $\vec{v_2}$ are vectors, $\cdot$ represents the dot product, and $\left \| \vec{v} \right \|$ the norm of $\vec{v}$. Its range is between $-1$ and $1$, where $1$ indicates that the vectors are in the same direction, $-1$ exactly the opposite direction, and $0$ indicates that vectors are orthogonal. It is worth mentioning that the absolute of the cosine similarity is used instead because a cosine-similarity value of $1$ or $-1$ is similar regarding the linear independence.

 \begin{equation}
CS(\vec{v_1},\vec{v_2}) = cos(\theta)  =  \frac{\vec{v_1} \cdot \vec{v_2}}{ \left \| \vec{v_1} \right \|  \left \| \vec{v_2} \right \|}
\label{eq:cosine_similarity}
\end{equation}

\begin{figure}[tb!]
\centering
\includegraphics[width=1.0\textwidth]{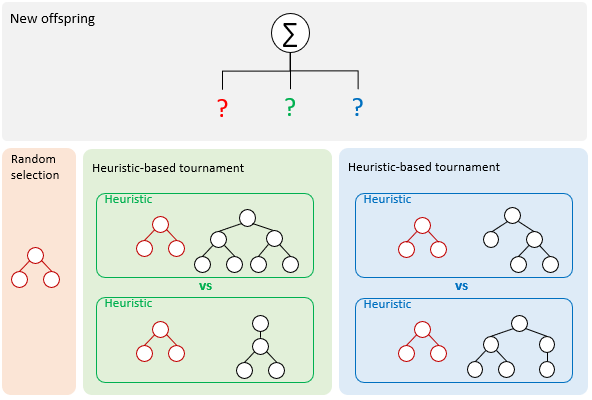}
 \caption[Diagram of parent selection based on heuristics]{ Diagram of parent selection based on heuristics.} 
  \label{fig:parent_selection_heu}
\end{figure}

The process of selecting a parent using the {\bf absolute cosine similarity} is depicted in Figure \ref{fig:parent_selection_heu}. Let us recall that tournament selection (with a tournament size of two) is being used, and the selection heuristic replaces the fitness function used traditionally in the tournament to select the parents. Under this configuration, the figure depicts the selection of three arguments for being used with the addition function. The first of the arguments is selected randomly from the population; this is depicted on the red box. Selecting the second argument (box in green) requires choosing two individuals from the population randomly and then comparing them using the absolute cosine similarity between each of the selected individuals and the first argument. The second argument is the one with the lowest value: the closest to a 90-degree angle. The absolute cosine similarity is obtained using the individuals' semantics, and these are vectors or a list of vectors (in case of multi-class problems); in the latter case, the average of the absolute cosine similarity is used instead. Finally, selecting the last argument (blue box) is equivalent to the previous one. That is, the individuals selected from the population are compared using the absolute cosine similarity with respect to the first argument selected. Although this process does not guarantee that all the arguments are unique, the implementation ensures that all individuals are different; this is depicted in the figure by representing each possible argument with a different tree. 

The second heuristic uses the {\bf Person's Correlation Coefficient} to select uncorrelated inputs. The correlation coefficient is defined in Equation \ref{eq:pearson_correlation_coefficient_sample}, where $\vec{v_1}$ and $\vec{v_2}$ are vectors with the values of the variables, $\cdot$ represents the dot product, $\bar{\vec{v}}$ is the average value of vector $\vec{v}$, and $\left \| \vec{v} \right \|$ the norm of $\vec{v}$. Pearson's range is between $-1$ and $1$, where $1$ indicates positively correlated, $0$ represents no linear correlation, and $-1$ represents a total negative linear correlation. It can be observed that Equations \ref{eq:cosine_similarity} and \ref{eq:pearson_correlation_coefficient_sample} are similar. The difference is that correlation (Equation \ref{eq:pearson_correlation_coefficient_sample}) subtracts the average value to the vectors. It means that the heuristics based on cosine similarity and correlation will be the same when the data is zero-centering.

\begin{equation}
 \rho_{\vec{v_1},\vec{v_2}} =\frac{(\vec{v_1}-\bar{\vec{v_1}}) \cdot (\vec{v_2}-\bar{\vec{v_2}})}{ \left \| \vec{v_1}-\bar{\vec{v_1}} \right \|  \left \| \vec{v_2}-\bar{\vec{v_2}} \right \|}
 \label{eq:pearson_correlation_coefficient_sample}
\end{equation}

Figure \ref{fig:parent_selection_heu} depicts the process of selecting three arguments for the addition function. The process is similar to the one used for the cosine similarity, being the only difference between them the use of the absolute value of the Pearson's coefficient instead of the cosine similarity. That is, the first argument is selected randomly from the population, whereas the second and third arguments are the individuals whose semantics obtained the lowest absolute Pearson's correlation coefficient for each tournament.

The previous selection heuristics increase the variety of the inputs using cosine similarity and Pearson's Correlation coefficient. However, these similarities do not consider the prediction labels. In classification problems, the individual's outputs are transformed to obtain the labels taking, for example, the maximum value index in a multiple output representation. The idea of the last heuristic is to complement the previous selection heuristics by measuring diversity using the predicted labels; the measure used with this purpose is named {\bf agreement} -- defined in the Equation \ref{eq:agreement_parents}\footnote{Note that in the case $\vec{p_2}$ is the target behavior, then the agreement is computing the accuracy of $\vec{p_1}$.}, where $\vec{p_1}$ and $\vec{p_2}$ represent the labels vectors of two individuals, $n$ is the number of samples, and $\delta(.)$ returns $1$ if its input is true and $0$ otherwise.

\begin{equation}
agr(\vec{p_1},\vec{p_2})=\frac{1}{n} \sum_i \delta(p_{1i} == p_{2i})
    \label{eq:agreement_parents}
\end{equation}

Figure \ref{fig:parent_selection_heu} depicts the procedure to select three arguments for the addition function using the agreement as the selection heuristic. The process is similar to the ones used for the previous selection heuristics. That is, the first argument is an individual randomly selected from the population. The second and third argument is selected performing a tournament where the fitness function is replaced by agreement. For example, the second argument (green box) is selected by first transforming the first argument's outputs into labels and transforming the outputs of the two individuals selected in the tournament. The labels obtained are used to compute two agreement values, one for each individual in the tournament using in common the labels of the first argument, i.e., the individual selected randomly. The second argument selected is the individual with the lowest agreement, i.e., the one that optimizes the variety. The third argument is selected performing another tournament following an equivalent method used on the second argument. 

To sum up, three selection heuristics are proposed, in this contribution,  corresponding to the use of the absolute cosine similarity, Pearsons' Correlation coefficient, and agreement. These heuristics replace the fitness function in the tournament selection procedure, and, consequently, the selected individuals are the ones with the lowest values on the particular heuristic used. 


\section{Experiments and results}
\label{sec:results}

The selection heuristics performance is analyzed in this section and compared against our GP system with the default parameters, with state-of-the-art selection heuristics, and with traditional classifiers and classifiers using full-model selection.

\subsection{Datasets}

The classification problems used as benchmarks are 30 datasets taken from the UCI repository \citep{DuaDheeruandGraff2017UCIRepository}. Table ~\ref{tab:dataset} shows the dataset information. It can be seen that the datasets are heterogeneous in terms of the number of samples, variables, and classes. Additionally, some of the classification problems are balanced, and others are imbalanced. We use Shannon's entropy to indicate the degree of the class-imbalance in the problem. It is defined as $H(X) = - \sum_i p_i \log(p_i) $, where $p_i$ represents the probability of the category $i$. We calculate those probabilities by counting the frequencies of each category. Besides, for normalizing, we use the logarithm base on the number of categories. For example, if the classification problem has four categories, we calculate the Shannon's entropy as $H(X) = - \sum_i p_i \log_4(p_i) $. In this sense, if the value is equal to $1.0$, it indicates a perfect balance problem. In opposite, the smaller the value, the bigger the imbalance.

\begin{table}[htbp!]
\centering
\caption[Datasets used to compare the performance of the algorithms]{Datasets used to compare the performance of the algorithms. These problems are taken from the UCI repository. The table includes Shannon's entropy to indicate the degree of the class-imbalance, where the value $1.0$ indicates that the samples are perfectly balanced. In opposite, the smaller the value, the bigger the imbalance.}
\begin{tabular}{|l||c|c|c|c|c|}
\hline
Dataset & Train   & Test    & Variables & Classes & Classes \\
        & samples & samples &           &         & entropy \\ \hline \hline
ad & 2295& 984 & 1557&2 & 0.58\\ \hline
adult & 32561& 16281 & 14&2 & 0.8\\ \hline
agaricus-lepiota & 5686 & 2438 & 22&7 & 0.81\\ \hline
aps-failure & 60000 & 16000 & 170&2 & 0.12\\ \hline
banknote & 960& 412 & 4&2 & 0.99\\ \hline
bank & 31647& 13564 & 16&2 & 0.52\\ \hline
biodeg & 738& 317 & 41&2 & 0.91\\ \hline
car & 1209& 519 & 6&4 & 0.6\\ \hline
census-income & 199523& 99762 & 41&2 & 0.34\\ \hline
cmc & 1031& 442 & 9&3 & 0.98\\ \hline
dota2 & 92650& 10294 & 116&2 & 1.0\\ \hline
drug-consumption & 1319& 566 & 30&7 & 0.44\\ \hline
fertility & 69& 30 & 9&2 & 0.43\\ \hline
IndianLiverPatient & 407& 175 & 10&2 & 0.85\\ \hline
iris & 105 & 45 & 4&3 & 1.0\\ \hline
krkopt & 19639& 8417 & 6&18 & 0.84\\ \hline
letter-recognition & 14000& 6000 & 16&26 & 1.0\\ \hline
magic04 & 13314& 5706 & 10&2 & 0.93\\ \hline
ml-prove & 4588& 1530 & 56&2 & 0.98\\ \hline
musk1 & 333& 143 & 166&2 & 0.99\\ \hline
musk2 & 4618& 1980 & 166&2 & 0.61\\ \hline
optdigits & 3823& 1797 & 64&10 & 1.0\\ \hline
page-blocks & 3831& 1642 & 10&5 & 0.27\\ \hline
parkinsons & 135& 59 & 22&2 & 0.79\\ \hline
pendigits & 7494& 3498 & 16&10 & 1.0\\ \hline
segmentation & 210& 2100 & 19&7 & 1.0\\ \hline
sensorless & 40956& 17553 & 48&11 & 1.0\\ \hline
tae & 105& 45 & 5&3 & 0.99\\ \hline
wine & 123& 53 & 13&3 & 0.99\\ \hline
yeast & 1038& 446 & 9&10 & 0.76\\ \hline
\end{tabular}
\label{tab:dataset}
\end{table}

The performance of the classifiers is measured in a test set. Some of the problems are already split between a training set and a test set in the repository. For those problems that this partition is not present, we performed cross-validation; that is, we randomly split the dataset using 70\% of the samples for the training set and 30\% for the test set. 

\subsection{Computer Equipment}
The computer characteristics where the experiments were executed are shown in Table \ref{tab:exp_computer_characteristics}. For a fair comparison, the experiments were executed using only one core.

\begin{table}[htbp!]
\centering
\caption{Characteristics of the computer where the experiments were executed}
\begin{tabular}{|c|c|}
\hline
Operating system & Ubuntu 16.04.2 LTS \\ \hline
Processor (CPU) & Intel (R) Xeon(R) CPU E5-2680 v4\\ \hline
Processor (CPU) speed & 2.5GHz \\ \hline
Computer memory size & 256 GB \\ \hline
Hard disk size & 1 TB \\ \hline
Cores number & 14 \\ \hline
\end{tabular}
\label{tab:exp_computer_characteristics}
\end{table}

\subsection{Performance Metrics}

The classifiers' performance is analyzed in terms of precision and time spent on training using two metrics: macro-F1 and time (in seconds) per sample.

Accuracy is maybe the most used metric for measuring the performance of classifiers. Its value ranges from $0$ to $1$, being one the best performance and zero the worst. It can be seen as the percentage of samples that are correctly predicted. However, if the classes are imbalanced, as the problems in this benchmark, accuracy is not reliable. On the other hand, the F1 score measures a binary classifier's performance taking into account the positive class. It is robust to imbalanced problems. For a multi-class problem, the F1 score can be extended as the macro-F1 score that corresponds to the average of the F1 score per class. Besides, most of the comparisons are performed based on the rank of macro-F1. It means, for each dataset, the classifiers are ranked according to their performance. The number 1 is assigned to the classifier with the highest value in macro-F1, number 2 corresponds to the one with the second-highest value in macro-F1, and so on. If several classifiers have the same value in macro-F1, they got the same rank, and the following rank number will be increased the number of repeated values. 

In addition to a classifier's performance for predicting the samples correctly, time is an essential factor in an algorithm. When the number of samples in the training set is small, all the algorithms learn the model quickly. However, if the number of samples grows, some algorithms spend considerably more and more time, and in some cases, it could be impossible to wait until the algorithm converges. As we mention in the previous section, the datasets vary on the number of samples, and, logically, algorithms spend more time learning big datasets. In that sense, to normalize the time, and with the idea of making comparisons based on this measure, we divided the time (in seconds) that algorithms spend in the training phase by the number of samples in the datasets.

\subsection{Comparison of the Proposed Selection Heuristics against Classic Tournament Selection}

We performed a comparison of different selection schemes for parent and negative selection. Specifically for {\bf parent selection}, we compare the use of the following techniques: (1) traditional tournament selection, which uses the individual's fitness ({\em fit}), (2) random selection ({\em rnd}), (3) tournament selection with the absolute of cosine similarity ({\em sim}), (4) tournament selection with the absolute of Pearson's correlation coefficient ({\em prs}), and, (5) tournament selection with the agreement ({\em agr}). The latest three selection methods correspond to the proposed selection heuristics; in this case, the selection heuristics are applied only for the functions that inspired them, i.e., addition ($\sum$), Naive Bayes ($\NB$ and $\MN$), and Nearest Centroid ($\NC$), in the rest of the functions random selection is used instead. In addition, for {\bf negative selection}, we analyze the use of the traditional negative selection, which uses the individual's fitness to select the worst individual in the tournament ({\em fit}), and random selection ({\em rnd}).

The selection schemes were tested on our GP system (EvoDAG). To improve the reading, we use the following notation. The selection scheme used for parent selection is followed by the symbol \say{-}, and then comes the abbreviation of the negative selection scheme. For example, sim-fit means tournament selection with the absolute cosine similarity for parent selection and traditional negative selection are used. In total, we analyze the performance of eight combinations: fit-fit, rnd-rnd, sim-fit, sim-rnd, prs-fit, prs-rnd, agr-fit, and agr-rnd. Furthermore, to complete the picture of the proposed selection heuristics and the relation with the functions that served as inspiration, we decided to include in the comparison the performance GP systems when the heuristics are used in all the functions with two or more arguments. These systems are identified with the symbol *. For example, agr-rnd indicates that agreement is used with the functions $\sum$, $\NB$, $\MN$, and $\NC$; whereas, agr-rnd* means that the heuristic is used with the functions: $\sum$, $\prod$, $\max$, $\min$, $\hypot$, $\NB$, $\MN$, and $\NC$. 

\begin{figure}
\centering
\begin{subfigure}[b]{0.4\textwidth}
    \centering
    \includegraphics[width=1.0\textwidth]{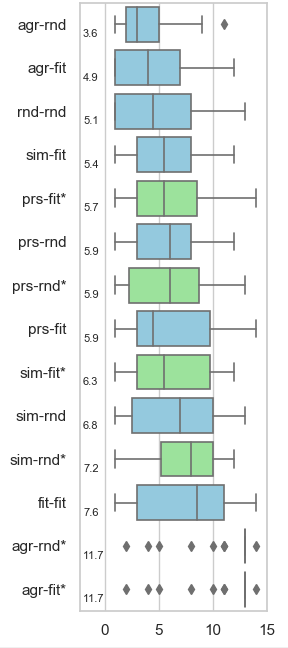}
    \caption{}
    \label{fig:res_selectionschemes_a}
\end{subfigure}
\begin{subfigure}[b]{0.4\textwidth}
    \centering
    \includegraphics[width=1.0\textwidth]{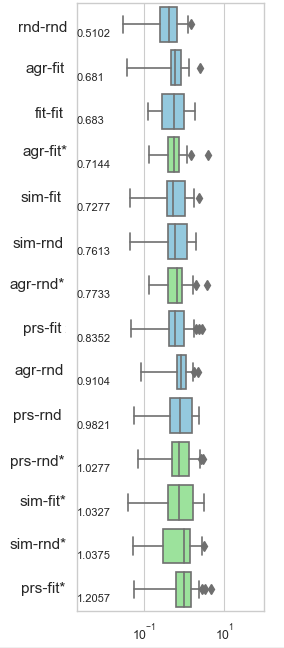}
    \caption{}
    \label{fig:res_selectionschemes_b}
\end{subfigure}
\caption{Selection schemes comparison. a) Macro-F1 ranks that are measured over the test datasets. b) Time, in seconds, required by the different selection technique combinations in the training phase. The time is divided by the number of train samples in the datasets. In both figures, green boxplots represent where the selection heuristics (sim, prs, or agr) are applied to all functions. The average rank, or time per sample, sorts the classifiers, and it appears on the left.}
\label{fig:res_selectionschemes}
\end{figure}

Figure \ref{fig:res_selectionschemes} shows the performance for the different techniques used for parent and negative selection on classification tasks. The detailed results can be observed on Table \ref{tab:res_apx_heuristics}. Figure \ref{fig:res_selectionschemes_a} shows the performance results based on macro-F1 ranks over the test sets. It can be seen that the best performance is obtained by selecting the parents with the agreement heuristic and random negative selection (agr-rnd), followed by the use of the same scheme for parent selection and negative tournament selection (agr-fit). The combinations agr-rnd, agr-fit, sim-fit, prs-fit*, prs-rnd, prs-fit, prs-rnd*, sim-fit*, sim-rnd, and sim-rnd* are better than fit-fit (i.e., selection using the fitness function) in terms of macro-F1 average rank. It means that our proposed heuristics improve the performance of the classical selection schemes (fit-fit). Random selection for parent and negative selection, rnd-rnd, also improves the performance of EvoDAG using the classical selection schemes based on fitness (fit-fit). It indicates the importance of population diversity, as mentioned in Novelty Search \cite{Lehman2011AbandoningAlone}. Besides, as we mention in Section \ref{sec:evodag}, once the individual is created, the function parameters are optimized using OLS and the target semantics. Our heuristic based on agreement works well because it is quite similar to the Novelty Search implemented in \citep{Naredo2016EvolvingSearch}, but instead of improving diversity among all individuals in the population, it enhances the diversity among parents.    

From the figure, it can also be observed that there exists a tendency when the heuristics (identified with the symbol *) are applied to all functions with more than one argument ($\sum$, $\prod$, $\max$, $\min$, $\hypot$, $\NB$, $\MN$, and $\NC$); the results are worse than those systems that use the proposed heuristics on the functions that inspired them (i.e., addition, Naive Bayes and Nearest Centroid). It affirms the importance of generating heuristics that are specifically designed based on the functions' properties. Comparing by the time that the classifiers spend in the training phase (see Figure \ref{fig:res_selectionschemes}.b), it can be seen that rnd-rnd is the fastest; this is because it is the most straightforward. 

For the statistical analysis, we use the Friedman and Nemenyi tests \citep{Demsar2006StatisticalSets}. Macro-F1 ranks values were used for the Friedman test where there was rejected the null hypothesis, with a p-value of $4.39e-22$. Based on Nemenyi test, the groups of techniques that are not significantly different (at p=0.10) are: Group 1 (agr-rnd, agr-fit, rnd-rnd, sim-fit, prs-fit*, prs-rnd, prs-rnd*, prs-fit, sim-fit*, sim-rnd, and sim-rnd*), Group 2 (agr-fit, rnd-rnd, sim-fit, prs-fit*, prs-rnd, prs-rnd*, prs-fit, sim-fit*, sim-rnd, sim-rnd*, and fit-fit), and Group 3 (agr-rnd*, agr-fit*). It indicates that our proposed heuristic combination based on the agreement for parent selection and random negative selection (agr-rnd) performs statistically better than the classical tournament selection using fitness (fit-fit) because they belong to different groups, i.e., Group 1 and Group 2, respectively. The test also indicates that there is not enough evidence to differentiate the systems of Group 2 (except fit-fit) with system agr-rnd.  

\subsection{Comparison of the Proposed Selection Heuristics against State-of-the-Art Selection Schemes}

As we mention in Section \ref{sec:related_work}, there are selection heuristics related to this research. Consequently, we decided to compare our selection heuristics with the two most similar methods; these are Angle-Driven-Selection \citep{Chen2019ImprovingOperators} and Novelty Search \citep{Lehman2011AbandoningAlone}. In Angle-Driven Selection (ads), the first individual is selected using traditional tournament selection and then replaces the fitness function, in the tournament selection, with their relative angle in the error space. As can be seen, in Angle-Drive-Selection, the first individual is chosen using the fitness, whereas, in our proposal, it is selected randomly. Therefore, we decided to add another parameter to indicate whether the first individual is selected using the fitness (fit) or random (rnd). The combinations of selection techniques' notation is as follows. The symbol \say{-} follows the parent selection technique, then comes the abbreviation of the negative selection scheme, and, for our heuristics and ads, at the ending, after the symbols \say{--}, comes the abbreviation of the scheme to select the first individual. For example, agr-rnd--fit means that agreement is used for parent selection, the negative selection is performed randomly, and the first individual is selected using the fitness. 

\begin{figure}[htb!]
\centering
\includegraphics[width=0.5\textwidth]{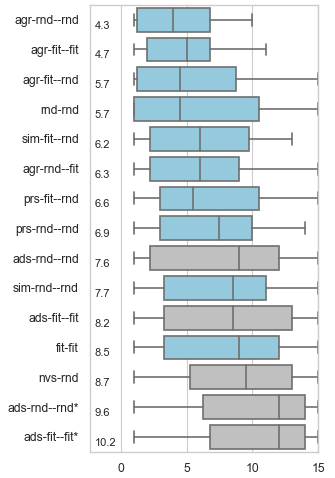} 
 \caption[Proposed heuristics against state-of-the-art selection schemes based on macro-F1]{ Proposed heuristics against state-of-the-art selection schemes based on macro-F1. Boxplots present the ranks, and those are measured using macro-F1 over the test datasets. Gray boxplots represent the selection techniques from the state-of-the-art, Novelty Search (nvs) and Angle-Driven Selection (ads). The average rank sorts the classifiers, and it appears on the left.} 
  \label{fig:res_selectiontechniques}
\end{figure} 

Figure \ref{fig:res_selectiontechniques} presents the results of the comparison, based on macro-F1, of our proposed heuristics, agreement, Pearson's Correlation coefficient, and cosine similarity (agr, prs, and sim) against state-of-the-art selection techniques: Angle-Driven Selection (ads) and Novelty Search (nvs). Tables \ref{tab:res_apx_selectionschemes_part1} and \ref{tab:res_apx_selectionschemes_part2} show the detailed results. It can be observed that the performance of our heuristics is generally better than Angle-Driven Selection (ads) and Novelty Search (nvs). Using Friedman and Nemenyi tests \citep{Demsar2006StatisticalSets}, it was found that agr-rnd--rnd, agr-fit--fit, agr-fit--rnd, rnd-rnd, sim-fit--rnd, agr-rnd--fit, prs-fit--rnd, prs-rnd--rnd, ads-rnd--rnd, and sim-rnd--rnd are not significantly different (at p = 0.10), but agr-rnd--rnd is significantly better than novelty search (nvs-rnd), angle-driven with the original proposal of selecting the first individual at random (ads-fit--fit), fit-fit, ads-rnd--rnd*, and ads-fit--fit*.

In the original proposal of Angle-Driven selection \citep{Chen2019ImprovingOperators}, it is implemented in a Geometric Semantic GP system. However, in this case, it is applied in EvoDAG. Angle-Driven selection is quite similar to the proposed heuristics based on Pearson's Correlation coefficient and cosine similarity (albeit following a different path). These techniques use the geometry of individuals' semantics for parent selection. ADS measures the angle between relative semantics (see Section \ref{sec:related_work}), while our heuristics measure the angle between the semantics and the centered semantics (see Section \ref{sec:selection}). On Figure \ref{fig:res_selectiontechniques}, it can be observed that Angle-Driven selection (ads-rnd--rnd) performs similar to Pearson's Correlation coefficient (prs-rnd--rnd) and cosine similarity (sim-rnd--rnd); in fact, its rank is in the middle of these two systems. Besides, Angle-Drive Selection is better when it uses the combination of selection schemes ads-rnd--rnd than the original proposal ads-fit--fit. As our heuristics behavior, we can see that Angle-Drive Selection works better when applied only to the functions addition, Naive Bayes, and Nearest Centroid than when applied to all functions with more than one argument.

On the other hand, Novelty Search was used in a traditional GP system to optimize the inputs of a Nearest Centroid classifier \citep{Naredo2016EvolvingSearch}. An individual's novelty is calculated from the whole population, not only of the individuals participating in the tournament as done in our proposed selection heuristics. Novelty Search's performance is just below our GP system with the default parameters (fit-fit), and it is the third system with the worst rank. The performance obtained by Novelty Search might indicate that it is better to use only the information of the individuals participating in the tournament to compute the similarity. The agreement selection heuristic could be seen as a way to transform the novelty search measure using only the individuals participating in the tournament. 

\subsection{Comparison of Proposed Selection Heuristics against State-of-the-Art Classifiers}

After analyzing the different selection schemes' performance, it is the moment to compare our selection heuristics against state-of-the-art classifiers. We chose the combination of the selection schemes: agr-rnd, rnd-rnd, fit-fit, ads-rnd, nvs-rnd. The reason is that agr-rnd is the combination that gives the best results, rnd-rnd represents the simplest schemes being also highly competitive, fit-fit represents the traditional tournament selection, and finally, ads-rnd and nvs-rnd are the state-of-the-art selection schemes. We decided to perform the comparison against sixteen classifiers of the scikit-learn python library \citep{Pedregosa2011Scikit-learn:Python},  all of them using their default parameters. Specifically, these classifiers are Perceptron, MLPClassifier, BernoulliNB, GaussianNB, KNeighborsClassifier, NearestCentroid, LogisticRegression, LinearSVC, SVC, SGDClassifier, PassiveAggressiveClassifier, DecisionTreeClassifier, ExtraTreesClassifier, RandomForestClassifier, AdaBoostClassifier and GradientBoostingClassifier. It is also included in the comparison two auto-machine learning libraries: autosklearn \citep{Feurer2015EfficientLearning} and TPOT \citep{Olson2016AutomatingOptimization}.

\begin{figure}[htbp!]
\centering
\includegraphics[width=0.6\textwidth]{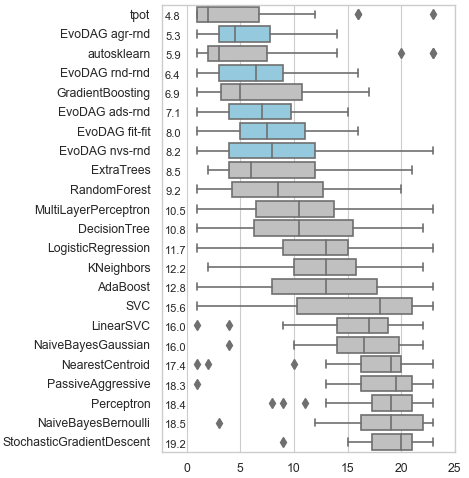}
 \caption [Comparison of EvoDAG against state-of-the-art classifiers based on macro-F1 rank]{Comparison of selection heuristics against state-of-the-art classifiers based on macro-F1 rank. The average rank sorts the classifiers, and those values appear on the left. The blue boxplots represent the selection heuristics.} 
  \label{fig:res_comparison_f1}
\end{figure}

\pagestyle{empty}
\begin{landscape}
\begin{table}[htbp!]
\supertiny
\centering
\caption[Comparison of EvoDAG against state-of-the-art classifiers based on macro-F1, the ranks are in parenthesis.]{Performance using macro-F1 (with ranks) of: tpot, autosklearn, Selection Heuristics (agr-rnd, rnd-rnd, fit-fit, nvs-rnd, and ads-rnd), Perceptron (PER), MLPClassifier (MLP), BernoulliNB (NBB), GaussianNB (NB), KNeighborsClassifier (KN), NearestCentroid (NC), LogisticRegression (LR), LinearSVC (LSVC), SVC, SGDClassifier (SDG), PassiveAggressiveClassifier (PA), DecisionTreeClassifier (DT), ExtraTreesClassifier (ET), RandomForestClassifier (RF), AdaBoostClassifier (AB) and GradientBoostingClassifier (GB). The symbol - represents that the classifier can not solve the classification problem. The rest of the systems are in Table \ref{tab:res_apx_classifiers_part2}.}
\begin{tabular}{ | l | r |  r |  r |  r |  r |  r |  r |  r |  r |  r |  r |  r |  }
\hline

& tpot & acc-rnd & autosklearn & rnd-rnd & GB & ads-rnd & fit-fit & nvs-rnd & ET & RF & MLP & DT \\ \hline \hline
ad&0.96(4) &0.94(8) &0.96(3) &0.93(14) &0.96(2) &0.96(5) &0.93(12) &0.94(7) &0.94(6) &0.96(1) &0.93(11) &0.94(10) \\ \hline 
adult&0.81(1) &0.79(4) &0.81(2) &0.79(5) &0.8(3) &0.79(9) &0.79(6) &0.79(8) &0.76(11) &0.77(10) &0.58(18) &0.74(12) \\ \hline 
agaricus-lepiota&0.67(7) &0.68(2) &0.68(5) &0.68(3) &0.56(16) &0.68(4) &0.68(6) &0.68(1) &0.43(21) &0.45(20) &0.61(11) &0.42(22) \\ \hline 
aps-failure&0.9(1) &0.83(8) &0.87(2) &0.86(3) &0.85(4) &0.84(6) &0.85(5) &0.83(9) &0.76(18) &0.8(12) &0.82(10) &0.83(7) \\ \hline 
banknote&1.0(1) &1.0(1) &1.0(10) &1.0(1) &0.99(17) &1.0(1) &1.0(1) &1.0(1) &1.0(12) &0.99(14) &1.0(1) &0.98(19) \\ \hline 
bank&0.74(6) &0.76(3) &0.71(8) &0.76(1) &0.72(7) &0.76(5) &0.76(4) &0.76(2) &0.68(12) &0.7(11) &0.67(13) &0.7(9) \\ \hline 
biodeg&0.81(8) &0.84(2) &0.82(6) &0.85(1) &0.79(12) &0.82(7) &0.83(5) &0.84(3) &0.81(10) &0.81(8) &0.8(11) &0.79(13) \\ \hline 
car&1.0(1) &0.87(6) &0.96(3) &0.86(7) &0.97(2) &0.8(11) &0.84(9) &0.81(10) &0.87(5) &0.85(8) &0.59(15) &0.94(4) \\ \hline 
census-income&0.78(1) &0.77(3) &0.75(6) &0.77(2) &0.75(5) &0.75(7) &0.75(4) &0.38(23) &0.72(9) &0.49(19) &0.71(10) &0.48(21) \\ \hline 
cmc&0.54(2) &0.54(4) &0.55(1) &0.53(8) &0.52(9) &0.53(5) &0.53(7) &0.53(6) &0.47(16) &0.48(14) &0.52(10) &0.45(18) \\ \hline 
dota2&0.59(7) &0.59(2) &0.59(8) &0.59(3) &0.56(13) &0.59(4) &0.59(1) &0.59(5) &0.55(15) &0.54(16) &0.59(10) &0.52(18) \\ \hline 
drug-consumption&0.18(12) &0.23(2) &0.13(20) &0.2(7) &0.19(11) &0.21(4) &0.2(8) &0.2(5) &0.16(16) &0.17(13) &0.25(1) &0.2(6) \\ \hline 
fertility&0.44(16) &0.45(4) &0.45(4) &0.45(4) &0.44(16) &0.45(4) &0.44(16) &0.45(4) &0.45(4) &0.45(4) &0.45(4) &0.59(2) \\ \hline 
IndianLiverPatient&0.55(16) &0.66(3) &0.56(14) &0.69(1) &0.55(15) &0.66(2) &0.64(5) &0.65(4) &0.61(6) &0.57(11) &0.59(8) &0.54(17) \\ \hline 
iris&0.98(8) &0.98(2) &0.98(2) &0.98(2) &0.94(14) &0.96(10) &0.98(2) &0.96(10) &0.92(17) &0.94(14) &0.98(2) &0.96(10) \\ \hline 
krkopt&0.91(1) &0.19(11) &-(23) &0.2(9) &0.61(6) &0.2(10) &0.15(15) &0.18(12) &0.7(4) &0.76(3) &0.54(8) &0.83(2) \\ \hline 
letter-recognition&0.97(2) &0.65(11) &-(23) &0.66(10) &0.91(7) &0.65(13) &0.65(13) &0.65(12) &0.94(4) &0.93(5) &0.92(6) &0.87(8) \\ \hline 
magic04&0.87(2) &0.85(5) &0.87(1) &0.85(6) &0.85(3) &0.84(9) &0.83(10) &0.84(8) &0.84(7) &0.85(4) &0.77(13) &0.79(12) \\ \hline 
ml-prove&1.0(1) &1.0(1) &1.0(1) &1.0(1) &1.0(1) &1.0(1) &1.0(13) &1.0(1) &1.0(17) &0.97(20) &1.0(13) &1.0(1) \\ \hline 
musk1&-(23) &0.88(5) &0.87(10) &0.87(11) &0.91(1) &0.89(4) &0.86(13) &0.9(3) &0.91(2) &0.87(9) &0.81(14) &0.79(16) \\ \hline 
musk2&0.98(1) &0.94(6) &0.98(2) &0.94(7) &0.93(10) &0.94(9) &0.95(3) &0.92(13) &0.95(4) &0.94(8) &0.95(5) &0.92(12) \\ \hline 
optdigits&0.98(1) &0.95(7) &0.98(2) &0.96(6) &0.96(4) &0.92(15) &0.94(11) &0.95(8) &0.95(9) &0.94(12) &0.96(5) &0.86(19) \\ \hline 
page-blocks&0.85(2) &0.83(4) &0.89(1) &0.76(12) &0.82(5) &0.76(11) &0.77(10) &0.78(9) &0.82(6) &0.79(7) &0.65(15) &0.85(3) \\ \hline 
parkinsons&0.81(5) &0.75(9) &0.82(3) &0.73(11) &0.85(1) &0.75(8) &0.67(15) &0.67(12) &0.84(2) &0.81(5) &0.43(18) &0.74(10) \\ \hline 
pendigits&0.98(1) &0.94(8) &0.98(3) &0.94(9) &0.96(6) &0.93(11) &0.94(10) &0.92(13) &0.96(5) &0.96(7) &0.97(4) &0.92(12) \\ \hline 
segmentation&0.95(1) &0.91(7) &0.94(2) &0.91(8) &0.94(4) &0.91(6) &0.9(11) &0.89(12) &0.93(5) &0.94(3) &0.68(16) &0.91(9) \\ \hline 
sensorless&1.0(1) &0.96(8) &1.0(3) &0.95(10) &0.99(5) &0.95(9) &0.96(7) &0.92(12) &1.0(2) &1.0(4) &0.95(11) &0.98(6) \\ \hline 
tae&0.51(5) &0.36(14) &0.45(6) &0.32(16) &0.52(4) &0.37(12) &0.44(7) &0.3(18) &0.54(3) &0.6(2) &0.4(8) &0.63(1) \\ \hline 
wine&1.0(1) &0.98(4) &1.0(1) &0.98(4) &0.98(4) &0.98(4) &0.98(4) &0.98(4) &0.98(4) &1.0(1) &0.14(23) &0.94(14) \\ \hline 
yeast&0.53(5) &0.47(6) &0.55(2) &0.45(9) &0.59(1) &0.46(7) &0.46(8) &0.45(10) &0.54(3) &0.44(12) &0.05(21) &0.45(11) \\ \hline  \hline
Average rank &4.8&5.3&5.9&6.4&6.9&7.1&8.0&8.2&8.5&9.2&10.5&10.8\\ \hline 

\end{tabular}
\label{tab:res_comparison_f1}
\end{table}
\end{landscape}

\begin{landscape}

\begin{figure*}[htbp!]
\centering
\includegraphics[width=1.5\textwidth]{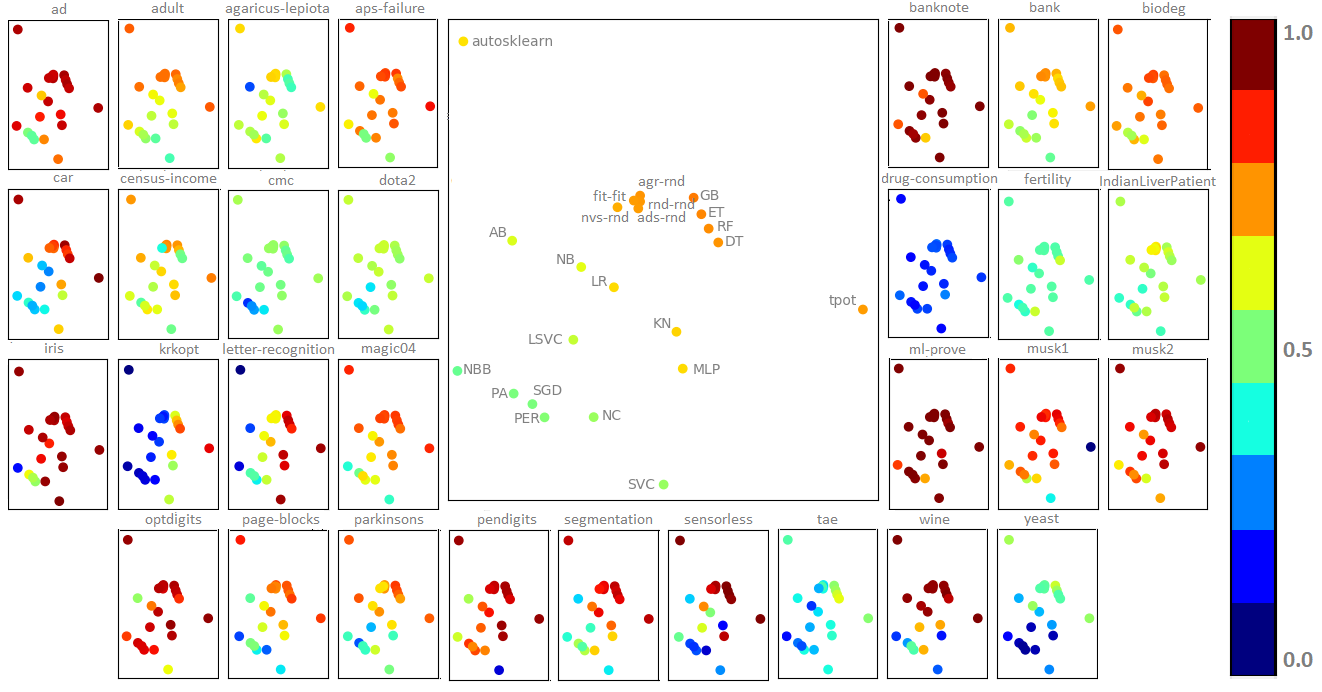}
 \caption [Performance of Selection Heuristics and state-of-the-art classifiers by dataset based on macro-F1]{Performance of Selection Heuristics and state-of-the-art classifiers by dataset based on macro-F1. The classifiers keep their position in all the images. Closer classifiers perform similarly. The macro-F1 value is represented by color, where dark red represents 1.0 and dark blue represents 0.0. The color scale is represented on the right. The systems are: tpot, autosklearn, Selection Heuristics (agr-rnd, rnd-rnd, fit-fit, nvs-rnd, and ads-rnd), Perceptron (PER), MLPClassifier (MLP), BernoulliNB (NBB), GaussianNB (NB), KNeighborsClassifier (KN), NearestCentroid (NC), LogisticRegression (LR), LinearSVC (LSVC), SVC, SGDClassifier (SDG), PassiveAggressiveClassifier (PA), DecisionTreeClassifier (DT), ExtraTreesClassifier (ET), RandomForestClassifier (RF), AdaBoostClassifier (AB) and GradientBoostingClassifier (GB)} 
  \label{fig:res_lpl_classifiers}
\end{figure*}

\end{landscape}
\pagestyle{plain}

Table \ref{tab:res_comparison_f1} and Figure \ref{fig:res_comparison_f1} show the comparison of classifiers based on macro-F1 ranks. The best classifier, based on the results of these experiments, is TPOT, followed by our GP system using agreement and random negative selection (agr-rnd), autosklearn, and GP with random selection (rnd-rnd) in both tournaments (positive and negative). It can be seen that the use of our proposed selection heuristic based on accuracy and negative random selection improves the performance of traditional selection (fit-fit) and positioned it into second place. The agreement selection heuristic with random negative selection produces a system that outperforms the scikit-learn classifiers, and it is competitive with auto-machine learning libraries, i.e., TPOT and autosklearn. Additionally, Table \ref{tab:res_comparison_f1} shows that autosklearn and TPOT cannot solve some classification problems. 

\begin{figure}[htbp!]
\centering
\includegraphics[width=1.0\textwidth]{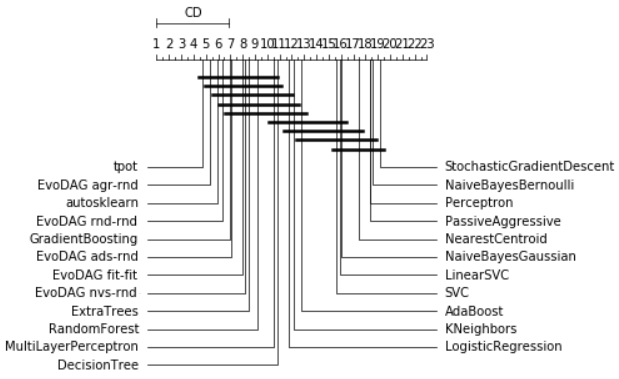}
 \caption [Comparison of all classifiers against each other using the macro-F1 average ranks with the Nemenyi test]{Comparison of all classifiers against each other using the macro-F1 average ranks with the Nemenyi test. Groups of classifiers that are not significantly different (at p = 0.10) are connected.} 
  \label{fig:res_comparison_classifiers_nemenyi_macrof1}
\end{figure}

Friedman's test rejects the null hypothesis that all classifiers perform similarly, with a p-value of $2.3e-54$. Nemenyi test (following the steps described in \cite{Demsar2006StatisticalSets}) results can be observed in Figure \ref{fig:res_comparison_classifiers_nemenyi_macrof1}. There were no statistical differences between TPOT, our proposed heuristics (agr-rnd, rnd-rnd, ads-rnd, nvs-rnd), autosklearn, GradientBoosting, ExtraTrees, RandomForest, and DecisionTree. It can also be observed that TPOT and agr-rnd belong to different groups (i.e., these are statistically different) than LogisticRegression, KNeighbors, AdaBoost, SVC, LinearSVC, Naive Bayes (Gaussian and Bernoulli), NearestCentroid, PassiveAggressive, Perceptron, and StochasticGradientDescent. 

In order to analyze the systems' performance, the difference in behavior, and the hardness of the problems, we decided to depict this information using our visualization technique proposed in \cite{sanchez2019liking}. The idea is to represent each classifier with a point in a plane. Each system can be seen as a vector where each dimension represents a problem, and the value is the system's performance (macro-F1) in that problem; using this representation, the idea is to depict this vector in a plane where the distance is preserved. Figure \ref{fig:res_lpl_classifiers} shows the classifiers' visualization; the color represents the macro-F1, and all the systems conserved the same position in all boxes. Small boxes represent a problem, and the one in the center is the average of all problems. The figure helps to identify those problems where all the systems behave similarly. For example, it can be observed that all the systems find easy the banknote and ml-prove problems. On the other hand, all the systems find complex the drug-consumption problem.  

From the figure, we can observe that the selection heuristics are close, the classifiers based on decision trees are close to the heuristics, and in opposite extremes are TPOT and autosklearn. Let us draw a line from the left upper corner to the right upper corner; we can see that the systems on the right of the line (including autosklearn) are in the top group shown in Figure \ref{fig:res_comparison_classifiers_nemenyi_macrof1} the only system missing is MLP which is on the left. 

\begin{figure}[htbp!]
\centering
\includegraphics[width=0.6\textwidth]{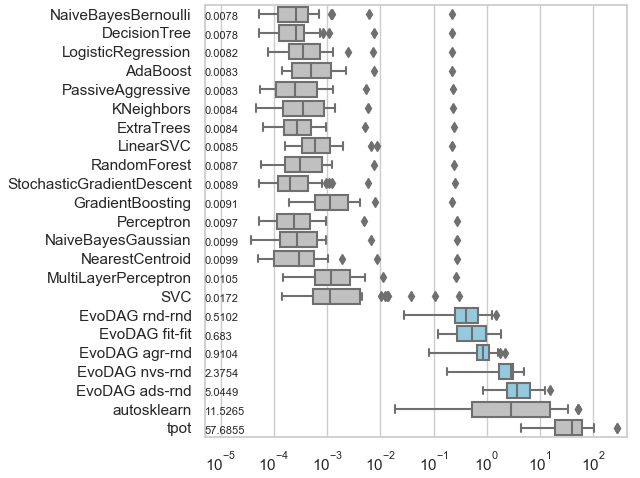}
 \caption[Comparison of EvoDAG against state-of-the-art classifiers based on the time required by the classifiers' training phase]{Comparison of selection heuristics against state-of-the-art classifiers based on the time required by the classifiers' training phase. The time is presented in seconds, and it is the average time per sample. The average time sorts the classifiers, and those values are on the left. The blue boxplots represent the selection heuristics. The time, represented on the x-axis, grows exponentially.} 
  \label{fig:res_comparison_time}
\end{figure}

\begin{figure}[htbp!]
\centering
\includegraphics[width=1.0\textwidth]{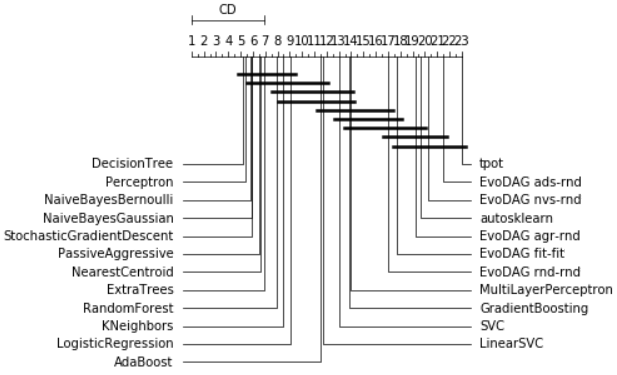}
 \caption [Comparison of all classifiers against each other using the time per sample average ranks with the Nemenyi test]{Comparison of all classifiers against each other using the time per sample average ranks with the Nemenyi test. Groups of classifiers that are not significantly different (at p = 0.10) are connected.} 
  \label{fig:res_comparison_classifiers_nemenyi_time}
\end{figure}

The classifiers' comparison based on the time spend in learning the model is presented in Figure \ref{fig:res_comparison_time}. Tables \ref{tab:res_apx_classifiers_time_part1}, \ref{tab:res_apx_classifiers_time_part2}, and \ref{tab:res_apx_classifiers_time_part3} show the detailed results. It can be seen that scikit-learn classifiers have the best ranks; these spend from 0.007 to 0.01 seconds per sample. With the different selection schemes, our GP system spends more time than scikit-learn classifiers in the learning phase. It spends, on average, from 0.5 to 5 seconds per sample. However, it is considerably faster than the auto-machine learning libraries, autosklearn and TPOT, which consume on average 11.5 and 57.68 seconds, respectively. Friedman's test rejects the null hypothesis that all classifiers spend the same time, with a p-value of $7.186e-94$. Nemenyi test (following \cite{Demsar2006StatisticalSets}) results can be observed in Figure \ref{fig:res_comparison_classifiers_nemenyi_time}. The figure shows a group formed by TPOT, the selection heuristics (except rnd-rnd), and autosklearn. The only selection heuristic that is statistically different from TPOT is rnd-rnd.

\section{Discussion}
\label{sec:discussion}

In this contribution, we have analyzed different selection schemes for GP.  The system used to perform the analysis is EvoDAG. So, it is essential to mention that the significant difference between EvoDAG and other GP systems is that each node contains constant(s) optimized using OLS to minimize the error. This characteristic is also present in TPOT and the Novelty Search Classifier \citep{Naredo2016EvolvingSearch}, albeit a different optimizer is used on those research works. The results obtained are in line with the results presented on Novelty Search \citep{Lehman2011AbandoningAlone,Naredo2016EvolvingSearch} where the idea is to abandon the fitness function, although the work presented by \cite{Naredo2016EvolvingSearch} and ours optimize constants in the evolutionary process. For those systems that do not optimize constants in the evolution, we believe the random selection schemes would not be competitive as these are in the current scenario. 

It is pertinent to mention the characteristics of the problems used as benchmarks, particularly the number of variables. It can be observed in Table \ref{tab:dataset} that the maximum number of variables is 1557, which corresponds to the "ad" dataset; also, this dataset is easy for the majority of classifiers (see Figure \ref{fig:res_lpl_classifiers}), and the second problem with more features is "aps-failure" with 170. Consequently, the analysis performed can guide selecting a classifier when the number of features is around 100, and it might not be of help regarding high-dimensional problems.  As a side note, in preliminary experiments, while doing research described in \cite{Graff2018EvoMSA:Analysis}, we did compare EvoDAG on text classification problems with a representation that contains more than 10 thousand features, and the result is that EvoDAG is not competitive against LinearSVC, neither in time nor in performance.   

Figure \ref{fig:res_lpl_classifiers} allows us to identify a limitation of our approach, let us look at the problems "krkopt" and letter-recognition; these problems are easily solved by TPOT and hard for GP using our selection heuristics. Although there is not enough evidence to draw some conclusions, it is observed that these problems are easily solved by classifiers based on Decision Trees (TPOT considers Decision Trees as its base learners) and contained the maximum number of classes, 18 and 26. As can be seen, Decision Trees are utterly different from the trees evolved by GP. Perhaps, the most distinctive characteristic is that the computation flow is the complement of GP trees; that is, the starting point is the root, and the output is a leaf. 

One GP characteristic that has captured researchers' attention is the ability to create white-box models; in this contribution, we have not adequately analyzed whether the models evolved by our GP system, using any selection heuristic, are easy or difficult to understand. Nonetheless, we have observed some of the models evolved in a few of the problems used as benchmarks. Our general impression is that the models evolved are complex containing at least ten inner nodes. For example, Figure \ref{fig:evodag_model} presents a model for the Iris problem; clearly, this problem could have been solved with a more straightforward tree obtaining similar performance. However, the system used did not promote the development of simple models, and we will address this issue in future work. 

\section{Conclusion}
\label{sec:conclusions}

In this research, we proposed three selection heuristics for parent selection in GP that used individuals' semantics and were inspired by functions' properties. These are described as follows. First, {\em tournament selection based on cosine similarity} (sim) aims to promote the selection of parents whose semantics' vectors ideally are orthogonal. {\em Tournament selection based on Pearson's Correlation coefficient} (prs) aims to promote the selection of parents whose semantics' vectors are uncorrelated. Finally, {\em tournament selection based on the agreement} (agr) tries to select parents whose predictions are different based on their prediction labels. These heuristics were inspired by the properties of the addition function, and the classifiers Naive Bayes and Nearest Centroid. To the best of our knowledge, this is the first time in Genetic Programming that functions' properties are taken into account to design methodologies for parent selection.

We compared our proposed heuristics against the classical parent selection technique, traditional tournament selection, and random parent selection. We also tested two state-of-the-art selection schemes, Novelty Search (nvs) and Angle-Driven Selection (ads). For negative selection, we tested the use of standard negative tournaments and random selection. Furthermore, our selection heuristics were compared against 18 state-of-the-art classifiers, 16 of them from the scikit-learn python library, and two auto-machine learning algorithms. The performance was analyzed on thirty classification problems taken from the UCI repository. The datasets were heterogeneous in terms of the number of samples, variables, and some of them are balanced, and others imbalanced.  

The results indicate that the selection heuristic using agreement combined with random negative selection (agr-rnd) is statistically better than the traditional selection that uses fitness (i.e., the system identified as fit-fit). On the other hand, on the comparison of the selection heuristics against different classifiers, it is observed that agr-rnd is a competitive classifier obtaining the second-best rank; additionally, the difference in performance with TPOT, which obtained the best rank, is not statistically significant.  Furthermore, it is observed that the selection heuristic identified as agr-rnd is grouped with the classifiers based on ensembles and the auto-machine learning algorithms; the group also includes Multilayer Perceptron and Decision Trees.

Finally, we have only tested our GP systems in classification problems and left aside regression problems. As can be observed, two of the selection heuristics develop, namely cosine similarity and Pearson's correlation, can be used without any modification on selection problems. On the other hand, the agreement heuristic is only defined for classification problems. We have performed some preliminary runs on regressions problems. The results indicated that the selection heuristics are competitive; however, we do not have enough evidence on whether these heuristics are different from traditional selection schemes or random selection on regression. We will deal with the comparison in regression problems as future work. 
\small

\bibliographystyle{apalike}
\bibliography{bibliography}

\pagestyle{empty}
\section{Appendix A}

This appendix contains all the detailed results. 

\begin{landscape}
\begin{table}[htbp!]
\supertiny
\centering
\caption{Selection schemes comparison based on macro-F1, the ranks are in parenthesis. Macro-f1 values were measured over test datasets.}
\begin{tabular}{ |l|r|r|r|r|r|r|r|r|r|r|r|r|r|r|r|r|r|r|r|r|r|r|r|r|r|r|r|r|r|r| }
\hline
 & acc-rnd & agr-fit & rnd-rnd & sim-fit & prs-fit* & prs-rnd & prs-rnd* & prs-fit & sim-fit* & sim-rnd & sim-rnd* & fit-fit & agr-rnd* & agr-fit* \\ \hline \hline
ad&0.94(1) &0.93(3) &0.93(8) &0.93(6) &0.93(10) &0.93(6) &0.92(12) &0.93(9) &0.93(3) &0.93(2) &0.92(11) &0.93(3) &0.55(13) &0.55(14) \\ \hline 
adult&0.79(2) &0.79(5) &0.79(3) &0.79(11) &0.79(12) &0.79(6) &0.79(4) &0.79(1) &0.79(7) &0.79(8) &0.79(9) &0.79(10) &0.69(13) &0.69(13) \\ \hline 
agaricus-lepiota&0.68(5) &0.68(3) &0.68(11) &0.68(4) &0.68(6) &0.68(8) &0.68(9) &0.68(1) &0.68(2) &0.68(10) &0.68(7) &0.68(12) &0.04(13) &0.04(13) \\ \hline 
aps-failure&0.83(11) &0.84(4) &0.86(1) &0.84(6) &0.84(7) &0.83(10) &0.85(2) &0.84(9) &0.84(5) &0.83(12) &0.84(8) &0.85(3) &0.74(13) &0.74(13) \\ \hline 
banknote&1.0(1) &1.0(1) &1.0(1) &1.0(1) &1.0(1) &1.0(1) &1.0(1) &1.0(1) &1.0(1) &1.0(1) &1.0(1) &1.0(1) &0.82(13) &0.82(13) \\ \hline 
bank&0.76(5) &0.76(8) &0.76(1) &0.76(2) &0.76(10) &0.76(4) &0.76(9) &0.76(3) &0.76(11) &0.76(7) &0.76(6) &0.76(12) &0.71(13) &0.71(13) \\ \hline 
biodeg&0.84(5) &0.84(7) &0.85(1) &0.84(3) &0.83(10) &0.82(12) &0.83(8) &0.85(2) &0.84(3) &0.84(6) &0.83(8) &0.83(11) &0.63(13) &0.63(13) \\ \hline 
car&0.87(5) &0.91(1) &0.86(6) &0.83(12) &0.87(3) &0.85(8) &0.86(7) &0.87(4) &0.89(2) &0.83(11) &0.84(10) &0.84(9) &0.29(13) &0.29(13) \\ \hline 
census-income&0.77(7) &0.51(12) &0.77(1) &0.77(3) &0.77(2) &0.77(4) &0.77(9) &0.51(11) &0.77(5) &0.77(8) &0.77(6) &0.75(10) &0.42(13) &0.42(13) \\ \hline 
cmc&0.54(9) &0.55(3) &0.53(12) &0.54(8) &0.54(7) &0.55(2) &0.55(1) &0.55(4) &0.55(5) &0.53(10) &0.54(6) &0.53(11) &0.47(13) &0.47(13) \\ \hline 
dota2&0.59(3) &0.59(7) &0.59(6) &0.6(1) &0.59(9) &0.59(12) &0.59(10) &0.59(11) &0.59(8) &0.59(4) &0.59(5) &0.59(2) &0.47(14) &0.48(13) \\ \hline 
drug-consumption&0.23(2) &0.2(10) &0.2(13) &0.22(3) &0.23(1) &0.21(8) &0.21(5) &0.21(9) &0.21(6) &0.22(4) &0.21(7) &0.2(14) &0.2(11) &0.2(11) \\ \hline 
fertility&0.45(1) &0.45(1) &0.45(1) &0.45(1) &0.45(1) &0.45(1) &0.45(1) &0.44(11) &0.45(1) &0.45(1) &0.45(1) &0.44(11) &0.4(13) &0.4(13) \\ \hline 
IndianLiverPatient&0.66(4) &0.71(1) &0.69(2) &0.65(8) &0.65(6) &0.64(12) &0.67(3) &0.65(6) &0.66(5) &0.64(10) &0.65(9) &0.64(11) &0.63(13) &0.63(13) \\ \hline 
iris&0.98(1) &0.98(1) &0.98(1) &0.98(1) &0.98(1) &0.98(1) &0.98(1) &0.96(10) &0.96(10) &0.98(1) &0.96(10) &0.98(1) &0.96(10) &0.96(10) \\ \hline 
krkopt&0.19(2) &0.18(4) &0.2(1) &0.14(11) &0.16(7) &0.16(6) &0.16(5) &0.18(3) &0.14(9) &0.14(12) &0.14(9) &0.15(8) &0.12(13) &0.12(13) \\ \hline 
letter-recognition&0.65(2) &0.65(4) &0.66(1) &0.65(5) &0.65(5) &0.65(5) &0.65(5) &0.65(3) &0.65(5) &0.65(5) &0.65(5) &0.65(5) &0.65(5) &0.65(5) \\ \hline 
magic04&0.85(4) &0.85(1) &0.85(7) &0.85(5) &0.84(9) &0.85(3) &0.84(8) &0.85(2) &0.84(11) &0.85(6) &0.84(10) &0.83(12) &0.65(13) &0.65(13) \\ \hline 
ml-prove&1.0(1) &1.0(1) &1.0(1) &1.0(1) &1.0(1) &1.0(1) &1.0(1) &1.0(1) &1.0(1) &1.0(1) &1.0(1) &1.0(12) &0.7(13) &0.7(13) \\ \hline 
musk1&0.88(5) &0.86(12) &0.87(9) &0.88(6) &0.89(3) &0.88(4) &0.88(7) &0.86(11) &0.87(8) &0.89(2) &0.9(1) &0.86(10) &0.37(13) &0.37(13) \\ \hline 
musk2&0.94(4) &0.91(12) &0.94(8) &0.94(6) &0.94(7) &0.95(3) &0.94(10) &0.94(5) &0.94(9) &0.95(2) &0.94(11) &0.95(1) &0.46(13) &0.46(13) \\ \hline 
optdigits&0.95(3) &0.95(6) &0.96(2) &0.95(8) &0.95(4) &0.95(7) &0.95(5) &0.96(1) &0.94(12) &0.95(9) &0.94(11) &0.94(10) &0.73(13) &0.73(13) \\ \hline 
page-blocks&0.83(1) &0.79(6) &0.76(12) &0.77(9) &0.8(4) &0.8(3) &0.81(2) &0.79(5) &0.76(11) &0.77(7) &0.77(10) &0.77(8) &0.76(13) &0.76(13) \\ \hline 
parkinsons&0.75(2) &0.75(1) &0.73(6) &0.73(3) &0.73(3) &0.72(7) &0.73(3) &0.65(14) &0.7(10) &0.72(7) &0.72(7) &0.67(13) &0.67(11) &0.67(11) \\ \hline 
pendigits&0.94(3) &0.95(2) &0.94(4) &0.93(9) &0.94(6) &0.95(1) &0.94(7) &0.94(5) &0.92(12) &0.93(10) &0.92(11) &0.94(8) &0.8(13) &0.8(13) \\ \hline 
segmentation&0.91(4) &0.9(6) &0.91(5) &0.9(10) &0.92(1) &0.9(8) &0.91(2) &0.91(3) &0.89(12) &0.9(9) &0.89(11) &0.9(7) &0.82(13) &0.82(13) \\ \hline 
sensorless&0.96(6) &0.97(1) &0.95(10) &0.95(9) &0.96(4) &0.96(5) &0.96(8) &0.96(3) &0.96(7) &0.95(12) &0.95(11) &0.96(2) &0.8(13) &0.8(13) \\ \hline 
tae&0.36(6) &0.32(12) &0.32(8) &0.42(3) &0.3(14) &0.32(7) &0.32(8) &0.42(4) &0.38(5) &0.3(13) &0.47(1) &0.44(2) &0.32(8) &0.32(8) \\ \hline 
wine&0.98(2) &0.98(2) &0.98(2) &0.98(2) &0.96(13) &0.98(11) &0.96(13) &0.98(11) &0.98(2) &0.98(2) &1.0(1) &0.98(2) &0.98(2) &0.98(2) \\ \hline 
yeast&0.47(1) &0.45(9) &0.45(8) &0.46(6) &0.46(3) &0.45(10) &0.45(11) &0.43(14) &0.46(2) &0.44(13) &0.44(12) &0.46(7) &0.46(4) &0.46(4) \\ \hline \hline
Average rank &3.6&4.9&5.1&5.4&5.7&5.9&5.9&5.9&6.3&6.8&7.2&7.6&11.7&11.7\\ \hline 
\end{tabular}
\label{tab:res_apx_heuristics}
\end{table}
\end{landscape}

\begin{landscape}
\begin{table}[htbp!]
\supertiny
\centering
\caption{Proposed heuristics against state-of-the-art selection schemes based on macro-F1, the ranks are in parenthesis. The table continues in Table \ref{tab:res_apx_selectionschemes_part2}.}
\begin{tabular}{  |l|r|r|r|r|r|r|r|r|r|r|r|r|r|r|r|r|r|r|r|r|r|r|r|r|r|r|}
\hline
 & agr-rnd--rnd & agr-fit--fit & agr-fit--rnd & rnd-rnd & sim-fit--rnd & agr-rnd--fit & prs-fit--rnd & prs-rnd--rnd & ads-rnd--rnd & sim-rnd--rnd & ads-fit--fit \\ \hline \hline
ad&0.94(5) &0.94(6) &0.93(9) &0.93(13) &0.93(11) &0.93(15) &0.93(14) &0.93(11) &0.96(1) &0.93(7) &0.93(8) \\ \hline 
adult&0.79(2) &0.79(11) &0.79(4) &0.79(3) &0.79(8) &0.79(9) &0.79(1) &0.79(5) &0.79(13) &0.79(6) &0.79(15) \\ \hline 
agaricus-lepiota&0.68(7) &0.69(2) &0.68(4) &0.68(11) &0.68(6) &0.69(1) &0.68(3) &0.68(9) &0.68(13) &0.68(10) &0.68(15) \\ \hline 
aps-failure&0.83(10) &0.84(7) &0.84(3) &0.86(1) &0.84(4) &0.83(8) &0.84(6) &0.83(9) &0.84(5) &0.83(13) &0.82(14) \\ \hline 
banknote&1.0(1) &1.0(1) &1.0(1) &1.0(1) &1.0(1) &1.0(1) &1.0(1) &1.0(1) &1.0(1) &1.0(1) &1.0(1) \\ \hline 
bank&0.76(8) &0.76(4) &0.76(10) &0.76(1) &0.76(5) &0.76(3) &0.76(6) &0.76(7) &0.76(12) &0.76(9) &0.75(14) \\ \hline 
biodeg&0.84(5) &0.85(2) &0.84(10) &0.85(1) &0.84(4) &0.84(9) &0.85(2) &0.82(14) &0.82(15) &0.84(8) &0.83(13) \\ \hline 
car&0.87(4) &0.86(5) &0.91(1) &0.86(7) &0.83(12) &0.87(2) &0.87(3) &0.85(8) &0.8(15) &0.83(11) &0.86(6) \\ \hline 
census-income&0.77(4) &0.76(6) &0.51(12) &0.77(1) &0.77(2) &0.51(13) &0.51(11) &0.77(3) &0.75(8) &0.77(5) &0.5(14) \\ \hline 
cmc&0.54(7) &0.53(11) &0.55(2) &0.53(15) &0.54(6) &0.53(12) &0.55(3) &0.55(1) &0.53(9) &0.53(10) &0.54(8) \\ \hline 
dota2&0.59(3) &0.59(10) &0.59(6) &0.59(5) &0.6(1) &0.59(13) &0.59(7) &0.59(11) &0.59(9) &0.59(4) &0.59(14) \\ \hline 
drug-consumption&0.23(1) &0.22(5) &0.2(12) &0.2(14) &0.22(2) &0.21(6) &0.21(9) &0.21(8) &0.21(10) &0.22(3) &0.2(13) \\ \hline 
fertility&0.45(1) &0.45(1) &0.45(1) &0.45(1) &0.45(1) &0.45(1) &0.44(12) &0.45(1) &0.45(1) &0.45(1) &0.45(1) \\ \hline 
IndianLiverPatient&0.66(7) &0.67(4) &0.71(1) &0.69(2) &0.65(9) &0.67(5) &0.65(8) &0.64(13) &0.66(6) &0.64(11) &0.68(3) \\ \hline 
iris&0.98(1) &0.98(1) &0.98(1) &0.98(1) &0.98(1) &0.98(1) &0.96(13) &0.98(1) &0.96(13) &0.98(1) &0.98(1) \\ \hline 
krkopt&0.19(5) &0.19(3) &0.18(8) &0.2(1) &0.14(13) &0.18(9) &0.18(6) &0.16(10) &0.2(2) &0.14(14) &0.19(4) \\ \hline 
letter-recognition&0.65(4) &0.66(2) &0.65(6) &0.66(1) &0.65(11) &0.66(3) &0.65(5) &0.65(11) &0.65(11) &0.65(11) &0.65(9) \\ \hline 
magic04&0.85(6) &0.85(1) &0.85(2) &0.85(9) &0.85(7) &0.85(4) &0.85(3) &0.85(5) &0.84(12) &0.85(8) &0.84(11) \\ \hline 
ml-prove&1.0(1) &1.0(1) &1.0(1) &1.0(1) &1.0(1) &1.0(1) &1.0(1) &1.0(1) &1.0(1) &1.0(1) &1.0(1) \\ \hline 
musk1&0.88(8) &0.88(5) &0.86(15) &0.87(12) &0.88(9) &0.88(10) &0.86(14) &0.88(7) &0.89(3) &0.89(2) &0.87(11) \\ \hline 
musk2&0.94(4) &0.94(8) &0.91(15) &0.94(7) &0.94(6) &0.93(10) &0.94(5) &0.95(3) &0.94(9) &0.95(2) &0.93(11) \\ \hline 
optdigits&0.95(3) &0.95(5) &0.95(7) &0.96(2) &0.95(9) &0.95(4) &0.96(1) &0.95(8) &0.92(12) &0.95(10) &0.92(13) \\ \hline 
page-blocks&0.83(1) &0.79(6) &0.79(5) &0.76(13) &0.77(10) &0.81(2) &0.79(4) &0.8(3) &0.76(11) &0.77(8) &0.76(12) \\ \hline 
parkinsons&0.75(6) &0.73(8) &0.75(4) &0.73(8) &0.73(7) &0.73(8) &0.65(15) &0.72(11) &0.75(4) &0.72(11) &0.76(2) \\ \hline 
pendigits&0.94(5) &0.95(3) &0.95(2) &0.94(6) &0.93(10) &0.94(4) &0.94(7) &0.95(1) &0.93(13) &0.93(11) &0.94(8) \\ \hline 
segmentation&0.91(3) &0.91(5) &0.9(8) &0.91(4) &0.9(13) &0.91(6) &0.91(2) &0.9(10) &0.91(1) &0.9(12) &0.91(7) \\ \hline 
sensorless&0.96(7) &0.97(2) &0.97(1) &0.95(12) &0.95(10) &0.96(8) &0.96(4) &0.96(6) &0.95(11) &0.95(14) &0.96(5) \\ \hline 
tae&0.36(8) &0.36(7) &0.32(13) &0.32(12) &0.42(3) &0.34(9) &0.42(4) &0.32(10) &0.37(5) &0.3(15) &0.44(1) \\ \hline 
wine&0.98(1) &0.98(1) &0.98(1) &0.98(1) &0.98(1) &0.98(1) &0.98(12) &0.98(12) &0.98(1) &0.98(1) &0.98(1) \\ \hline 
yeast&0.47(1) &0.45(8) &0.45(6) &0.45(5) &0.46(3) &0.44(12) &0.43(15) &0.45(7) &0.46(2) &0.44(11) &0.45(10) \\ \hline \hline 
Average rank &4.3&4.7&5.7&5.7&6.2&6.3&6.6&6.9&7.6&7.7&8.2\\ \hline 
\end{tabular}
\label{tab:res_apx_selectionschemes_part1}
\end{table}
\end{landscape}

\begin{landscape}
\begin{table}[htbp!]
\supertiny
\centering
\caption{Proposed heuristics against state-of-the-art selection schemes based on macro-F1, the ranks are in parenthesis.}
\begin{tabular}{  |l|r|r|r|r|}
\hline
 & fit-fit & nvs-rnd & ads-rnd--rnd* & ads-fit--fit* \\ \hline \hline 
ad&0.93(9) &0.94(4) &0.95(2) &0.94(3) \\ \hline 
adult&0.79(7) &0.79(10) &0.79(14) &0.79(12) \\ \hline 
agaricus-lepiota&0.68(14) &0.68(5) &0.68(8) &0.68(12) \\ \hline 
aps-failure&0.85(2) &0.83(11) &0.83(12) &0.81(15) \\ \hline 
banknote&1.0(1) &1.0(1) &1.0(1) &1.0(1) \\ \hline 
bank&0.76(11) &0.76(2) &0.75(15) &0.75(13) \\ \hline 
biodeg&0.83(11) &0.84(7) &0.83(12) &0.84(6) \\ \hline 
car&0.84(9) &0.81(14) &0.83(10) &0.81(13) \\ \hline 
census-income&0.75(7) &0.38(15) &0.74(9) &0.74(10) \\ \hline 
cmc&0.53(14) &0.53(13) &0.54(4) &0.54(5) \\ \hline 
dota2&0.59(2) &0.59(12) &0.59(8) &0.59(15) \\ \hline 
drug-consumption&0.2(15) &0.2(11) &0.21(7) &0.22(4) \\ \hline 
fertility&0.44(12) &0.45(1) &0.44(12) &0.44(12) \\ \hline 
IndianLiverPatient&0.64(12) &0.65(9) &0.62(15) &0.63(14) \\ \hline 
iris&0.98(1) &0.96(13) &0.98(1) &0.98(1) \\ \hline 
krkopt&0.15(11) &0.18(7) &0.15(12) &0.13(15) \\ \hline 
letter-recognition&0.65(11) &0.65(7) &0.65(8) &0.65(10) \\ \hline 
magic04&0.83(15) &0.84(10) &0.84(13) &0.84(14) \\ \hline 
ml-prove&1.0(14) &1.0(1) &1.0(1) &1.0(14) \\ \hline 
musk1&0.86(13) &0.9(1) &0.88(5) &0.89(4) \\ \hline 
musk2&0.95(1) &0.92(13) &0.91(14) &0.93(12) \\ \hline 
optdigits&0.94(11) &0.95(6) &0.92(14) &0.92(15) \\ \hline 
page-blocks&0.77(9) &0.78(7) &0.76(15) &0.76(14) \\ \hline 
parkinsons&0.67(14) &0.67(13) &0.76(2) &0.78(1) \\ \hline 
pendigits&0.94(9) &0.92(15) &0.93(14) &0.93(12) \\ \hline 
segmentation&0.9(9) &0.89(15) &0.89(14) &0.9(11) \\ \hline 
sensorless&0.96(3) &0.92(15) &0.95(13) &0.96(9) \\ \hline 
tae&0.44(2) &0.3(14) &0.37(6) &0.32(11) \\ \hline 
wine&0.98(1) &0.98(1) &0.96(14) &0.96(14) \\ \hline 
yeast&0.46(4) &0.45(9) &0.44(13) &0.43(14) \\ \hline 
Average rank &8.5&8.7&9.6&10.2\\ \hline 
\end{tabular}
\label{tab:res_apx_selectionschemes_part2}
\end{table}
\end{landscape}

\begin{landscape}
\begin{table}[htbp!]
\supertiny
\centering
\caption[Comparison of EvoDAG against state-of-the-art classifiers based on macro-F1, the ranks are in parenthesis.]{Performance using macro-F1 (with ranks) of: LogisticRegression (LR), LinearSVC (LSVC), SVC, SGDClassifier (SDG), PassiveAggressiveClassifier (PA), DecisionTreeClassifier (DT), ExtraTreesClassifier (ET), RandomForestClassifier (RF), AdaBoostClassifier (AB) and GradientBoostingClassifier (GB). The begining of the table appears on Table \ref{tab:res_comparison_f1}.}
\begin{tabular}{ | l | r |  r |  r |  r |  r |  r |  r |  r |  r |  r |  r |  r |  r |  r |  r |  r |  r |  r |   }
\hline

 & LR & KN & AB & SVC & LSVC & NB & NC & PA & PER & NBB & SGD \\ \hline \hline
ad&0.94(9) &0.9(16) &0.93(13) &0.79(18) &0.88(17) &0.7(20) &0.77(19) &0.5(21) &0.46(22) &0.92(15) &0.46(22) \\ \hline 
adult&0.64(15) &0.63(16) &0.79(7) &0.44(23) &0.58(19) &0.64(14) &0.46(22) &0.61(17) &0.52(21) &0.68(13) &0.55(20) \\ \hline 
agaricus-lepiota&0.62(10) &0.53(18) &0.2(23) &0.56(15) &0.64(9) &0.57(13) &0.46(19) &0.56(14) &0.66(8) &0.58(12) &0.55(17) \\ \hline 
aps-failure&0.79(14) &0.78(16) &0.8(11) &0.53(21) &0.79(13) &0.63(20) &0.77(17) &0.78(15) &0.49(23) &0.65(19) &0.49(22) \\ \hline 
banknote&0.99(14) &1.0(10) &1.0(1) &1.0(1) &0.99(16) &0.82(21) &0.69(23) &0.99(13) &0.98(18) &0.82(22) &0.97(20) \\ \hline 
bank&0.63(16) &0.65(15) &0.7(10) &0.47(23) &0.51(22) &0.67(14) &0.54(20) &0.53(21) &0.59(17) &0.59(18) &0.55(19) \\ \hline 
biodeg&0.84(3) &0.77(17) &0.78(15) &0.78(16) &0.79(14) &0.72(19) &0.62(21) &0.57(22) &0.72(20) &0.74(18) &0.55(23) \\ \hline 
car&0.26(23) &0.74(12) &0.71(13) &0.69(14) &0.26(22) &0.32(20) &0.37(17) &0.41(16) &0.32(19) &0.34(18) &0.3(21) \\ \hline 
census-income&0.68(11) &0.68(12) &0.73(8) &0.48(20) &0.56(18) &0.59(17) &0.63(14) &0.45(22) &0.66(13) &0.61(15) &0.61(16) \\ \hline 
cmc&0.48(13) &0.49(12) &0.5(11) &0.54(3) &0.48(15) &0.46(17) &0.35(20) &0.17(23) &0.28(21) &0.44(19) &0.26(22) \\ \hline 
dota2&0.59(6) &0.52(17) &0.58(11) &0.59(9) &0.35(21) &0.56(14) &0.5(19) &0.35(22) &0.41(20) &0.56(12) &0.34(23) \\ \hline 
drug-consumption&0.16(15) &0.16(14) &0.13(23) &0.13(21) &0.14(17) &0.14(18) &0.19(10) &0.14(19) &0.2(9) &0.23(3) &0.13(22) \\ \hline 
fertility&0.45(4) &0.45(4) &0.52(3) &0.45(4) &0.45(4) &0.41(21) &0.6(1) &0.38(23) &0.41(21) &0.44(16) &0.43(20) \\ \hline 
IndianLiverPatient&0.5(18) &0.57(12) &0.59(7) &0.43(20) &0.41(21) &0.57(10) &0.56(13) &0.41(22) &0.47(19) &0.41(22) &0.58(9) \\ \hline 
iris&0.88(19) &0.98(8) &0.94(14) &1.0(1) &0.9(18) &0.96(10) &0.98(2) &0.64(20) &0.53(22) &0.13(23) &0.56(21) \\ \hline 
krkopt&0.18(13) &0.66(5) &0.1(18) &0.58(7) &0.16(14) &0.13(16) &0.12(17) &0.05(21) &0.08(19) &0.04(22) &0.08(20) \\ \hline 
letter-recognition&0.71(9) &0.94(3) &0.19(21) &0.97(1) &0.6(16) &0.64(15) &0.59(17) &0.45(19) &0.36(20) &0.08(22) &0.45(18) \\ \hline 
magic04&0.75(15) &0.76(14) &0.82(11) &0.41(22) &0.66(17) &0.65(18) &0.63(20) &0.48(21) &0.63(19) &0.39(23) &0.67(16) \\ \hline 
ml-prove&1.0(1) &0.94(21) &1.0(1) &0.99(18) &1.0(1) &1.0(16) &0.73(23) &1.0(1) &0.99(19) &0.85(22) &1.0(15) \\ \hline 
musk1&0.88(6) &0.88(6) &0.88(8) &0.37(22) &0.86(12) &0.77(18) &0.68(20) &0.81(14) &0.57(21) &0.72(19) &0.78(17) \\ \hline 
musk2&0.91(14) &0.92(11) &0.9(15) &0.73(21) &0.9(16) &0.75(20) &0.61(23) &0.85(17) &0.83(18) &0.66(22) &0.76(19) \\ \hline 
optdigits&0.95(10) &0.98(3) &0.53(23) &0.64(22) &0.93(13) &0.79(21) &0.89(18) &0.93(14) &0.91(17) &0.84(20) &0.92(16) \\ \hline 
page-blocks&0.79(8) &0.71(13) &0.46(19) &0.35(21) &0.61(16) &0.65(14) &0.22(22) &0.5(18) &0.36(20) &0.19(23) &0.5(17) \\ \hline 
parkinsons&0.76(7) &0.67(14) &0.82(4) &0.49(17) &0.3(22) &0.67(12) &0.6(16) &0.21(23) &0.41(21) &0.43(18) &0.43(18) \\ \hline 
pendigits&0.89(14) &0.98(2) &0.55(22) &0.08(23) &0.81(18) &0.82(17) &0.77(19) &0.86(15) &0.84(16) &0.6(21) &0.77(20) \\ \hline 
segmentation&0.9(10) &0.8(13) &0.33(23) &0.36(22) &0.42(20) &0.79(14) &0.69(15) &0.56(18) &0.57(17) &0.4(21) &0.45(19) \\ \hline 
sensorless&0.5(15) &0.11(22) &0.33(17) &0.26(18) &0.61(14) &0.76(13) &0.07(23) &0.19(20) &0.2(19) &0.48(16) &0.18(21) \\ \hline 
tae&0.34(15) &0.38(10) &0.37(13) &0.38(9) &0.3(19) &0.17(22) &0.31(17) &0.24(20) &0.38(11) &0.13(23) &0.2(21) \\ \hline 
wine&0.98(12) &0.73(15) &0.98(4) &0.21(21) &0.71(17) &0.96(13) &0.72(16) &0.28(20) &0.44(19) &0.18(22) &0.46(18) \\ \hline 
yeast&0.31(13) &0.28(15) &0.3(14) &0.26(16) &0.05(18) &0.54(4) &0.05(20) &0.05(19) &0.03(22) &0.12(17) &0.02(23) \\ \hline \hline
Average rank &11.7&12.2&12.8&15.6&16.0&16.0&17.4&18.3&18.4&18.5&19.2\\ \hline 

\end{tabular}
\label{tab:res_apx_classifiers_part2}
\end{table}
\end{landscape}

\begin{landscape}
\begin{table}[htbp!]
\supertiny
\centering
\caption[Comparison of EvoDAG against state-of-the-art classifiers based on time, the ranks are in parenthesis.]{Performance using time (with ranks) of: LogisticRegression (LR), LinearSVC (LSVC), SVC, SGDClassifier (SDG), PassiveAggressiveClassifier (PA), DecisionTreeClassifier (DT), ExtraTreesClassifier (ET), RandomForestClassifier (RF), AdaBoostClassifier (AB) and GradientBoostingClassifier (GB). Part 1.}
\begin{tabular}{|l |r|r|r|r|r|r|r|r|}
\hline
 & NBB & DT & LR & AB & PA & KN & ET & LSVC \\ \hline \hline
ad&6.16e-03(6) &7.58e-03(10) &7.17e-03(8) &7.49e-03(9) &5.34e-03(3) &5.82e-03(5) &5.06e-03(2) &8.79e-03(14) \\ \hline 
adult&1.24e-04(3) &1.33e-04(6) &1.33e-04(7) &1.58e-04(11) &1.25e-04(4) &1.41e-04(9) &1.58e-04(10) &2.46e-04(15) \\ \hline 
agaricus-lepiota&5.03e-04(11) &2.75e-04(4) &5.45e-04(12) &4.77e-04(6) &4.92e-04(8) &2.84e-04(5) &2.72e-04(3) &6.6e-04(14) \\ \hline 
aps-failure&2.19e-01(5) &2.19e-01(4) &2.24e-01(9) &2.22e-01(7) &2.34e-01(11) &2.33e-01(10) &2.39e-01(12) &2.21e-01(6) \\ \hline 
banknote&5.53e-04(9) &4.86e-04(7) &2.39e-04(4) &4.16e-04(5) &7.93e-04(14) &1.79e-04(1) &4.63e-04(6) &6.67e-04(10) \\ \hline 
bank&1.19e-04(6) &1.43e-04(9) &1.30e-04(8) &1.47e-04(10) &1.06e-04(2) &1.60e-04(13) &1.52e-04(12) &2.45e-04(15) \\ \hline 
biodeg&2.66e-04(5) &1.24e-04(1) &7.29e-04(15) &7.27e-04(13) &2.49e-04(4) &6.82e-04(11) &2.70e-04(6) &3.34e-04(7) \\ \hline 
car&7.45e-05(5) &6.55e-05(2) &7.88e-05(6) &1.75e-04(13) &6.96e-05(3) &7.08e-05(4) &8.94e-05(8) &2.82e-04(14) \\ \hline 
census-income&3.72e-04(2) &3.74e-04(3) &5.14e-04(12) &5.11e-04(11) &4.14e-04(7) &1.39e-03(15) &3.69e-04(1) &7.54e-04(14) \\ \hline 
cmc&1.86e-04(6) &1.27e-04(3) &8.77e-04(12) &7.63e-04(10) &1.36e-04(4) &1.05e-03(15) &1.57e-04(5) &3.50e-04(8) \\ \hline 
dota2&4.43e-04(5) &5.23e-04(11) &4.28e-04(3) &4.95e-04(8) &4.21e-04(1) &1.13e-03(15) &4.99e-04(9) &7.27e-04(14) \\ \hline 
drug-consumption&2.47e-04(4) &2.7e-04(6) &3.68e-04(12) &3.86e-04(13) &2.89e-04(7) &3.05e-04(9) &3.07e-04(10) &7.5e-04(14) \\ \hline 
fertility&2.57e-04(6) &1.47e-04(1) &1.82e-04(2) &1.59e-03(15) &1.89e-04(4) &5.13e-04(8) &6.62e-04(12) &1.86e-04(3) \\ \hline 
IndianLiverPatient&1.94e-04(8) &2.48e-04(12) &1.05e-04(5) &4.38e-04(15) &5.56e-05(1) &1.46e-04(7) &2.10e-04(9) &4.21e-04(14) \\ \hline 
iris&3.76e-04(7) &8.10e-05(1) &3.48e-04(6) &1.87e-03(13) &1.52e-04(4) &5.32e-04(8) &1.58e-04(5) &1.95e-03(16) \\ \hline 
krkopt&6.54e-05(4) &5.31e-05(2) &1.21e-04(11) &1.48e-04(12) &7.25e-05(10) &6.8e-05(7) &6.64e-05(6) &1.21e-03(13) \\ \hline 
letter-recognition&5.26e-05(1) &5.95e-05(3) &4.87e-04(12) &2.31e-04(11) &9.84e-05(9) &1.93e-04(10) &8.62e-05(6) &2.01e-03(14) \\ \hline 
magic04&7.55e-05(7) &8.07e-05(10) &7.87e-05(9) &1.65e-04(14) &7.82e-05(8) &7.39e-05(6) &7.14e-05(5) &1.64e-04(13) \\ \hline 
ml-prove&2.63e-04(5) &2.75e-04(8) &2.15e-04(2) &5.17e-04(14) &2.41e-04(4) &3.87e-04(11) &2.67e-04(6) &4.67e-04(13) \\ \hline 
musk1&1.20e-03(8) &1.08e-03(7) &1.31e-03(12) &2.24e-03(14) &1.28e-03(11) &1.03e-03(4) &8.95e-04(2) &1.43e-03(13) \\ \hline 
musk2&6.9e-04(2) &7.28e-04(3) &1.22e-03(12) &1.45e-03(13) &8.74e-04(5) &1.21e-03(11) &6.25e-04(1) &8.15e-04(4) \\ \hline 
optdigits&3.51e-04(4) &3.64e-04(6) &7.34e-04(14) &4.47e-04(10) &1.92e-04(1) &5.26e-04(12) &3.87e-04(9) &5.22e-04(11) \\ \hline 
page-blocks&7.74e-05(9) &6.02e-05(5) &1.66e-04(12) &1.4e-04(11) &7.09e-05(8) &4.50e-05(2) &6.11e-05(6) &3.38e-04(13) \\ \hline 
parkinsons&2.35e-04(3) &1.65e-04(1) &2.11e-04(2) &1.03e-03(16) &3.41e-04(6) &7.01e-04(13) &4.48e-04(11) &2.96e-04(4) \\ \hline 
pendigits&9.51e-05(4) &9.49e-05(3) &3.25e-04(13) &1.9e-04(11) &9.90e-05(6) &1.34e-04(10) &9.83e-05(5) &3.53e-04(14) \\ \hline 
segmentation&1.22e-03(12) &8.18e-04(4) &1.05e-03(9) &1.25e-03(14) &1.13e-03(10) &9.49e-04(6) &9.67e-04(8) &1.23e-03(13) \\ \hline 
sensorless&2.60e-04(6) &3.32e-04(9) &2.46e-03(13) &5.18e-04(11) &2.74e-04(7) &4.88e-04(10) &1.51e-04(2) &6.67e-03(15) \\ \hline 
tae&3.98e-04(10) &2.64e-04(4) &4.72e-04(11) &1.38e-03(14) &6.79e-04(12) &2.76e-04(6) &3.68e-04(9) &3.5e-04(8) \\ \hline 
wine&1.18e-04(1) &3.02e-04(7) &3.37e-04(8) &1.02e-03(15) &6.90e-04(14) &1.31e-04(4) &6.17e-04(13) &4.03e-04(10) \\ \hline 
yeast&2.62e-04(11) &2.02e-04(4) &2.69e-04(12) &2.16e-04(7) &2.06e-04(6) &2.05e-04(5) &2.44e-04(9) &8.52e-04(15) \\ \hline \hline
Average time per sample &0.0078&0.0078&0.0082&0.0083&0.0083&0.0084&0.0084&0.0085\\ \hline 
Average rank &5.8&5.2&9.1&11.5&6.5&8.4&6.9&11.7\\ \hline 
\end{tabular}
\label{tab:res_apx_classifiers_time_part1}
\end{table}
\end{landscape}

\begin{landscape}
\begin{table}[htbp!]
\supertiny
\centering
\caption[Comparison of EvoDAG against state-of-the-art classifiers based on time, the ranks are in parenthesis.]{Performance using time (with ranks) of: LogisticRegression (LR), LinearSVC (LSVC), SVC, SGDClassifier (SDG), PassiveAggressiveClassifier (PA), DecisionTreeClassifier (DT), ExtraTreesClassifier (ET), RandomForestClassifier (RF), AdaBoostClassifier (AB) and GradientBoostingClassifier (GB). Part 2.}
\begin{tabular}{| l |r|r|r|r|r|r|r|r|}
\hline
& RF & SGD & GB & PER & NB & NC & MLP & SVC \\ \hline \hline
ad&7.62e-03(11) &5.78e-03(4) &7.76e-03(12) &4.97e-03(1) &6.58e-03(7) &8.48e-03(13) &1.10e-02(16) &1.03e-02(15) \\ \hline 
adult&1.58e-04(12) &1.23e-04(2) &2.17e-04(13) &1.31e-04(5) &1.40e-04(8) &1.17e-04(1) &2.29e-04(14) &1.38e-02(16) \\ \hline 
agaricus-lepiota&2.71e-04(1) &4.87e-04(7) &1.07e-03(16) &4.97e-04(9) &2.71e-04(2) &4.97e-04(10) &9.10e-04(15) &6.49e-04(13) \\ \hline 
aps-failure&2.43e-01(13) &2.52e-01(14) &2.22e-01(8) &2.78e-01(16) &2.8e-01(18) &2.78e-01(17) &2.62e-01(15) &2.96e-01(19) \\ \hline 
banknote&7.02e-04(11) &7.97e-04(15) &5.26e-04(8) &7.45e-04(12) &1.81e-04(3) &1.80e-04(2) &3.11e-03(16) &7.53e-04(13) \\ \hline 
bank&1.2e-04(7) &1.17e-04(5) &1.95e-04(14) &1.15e-04(4) &1.01e-04(1) &1.10e-04(3) &1.48e-04(11) &1.30e-02(16) \\ \hline 
biodeg&6.77e-04(9) &2.36e-04(3) &5.86e-04(8) &2.36e-04(2) &6.83e-04(12) &7.28e-04(14) &1.53e-03(16) &6.82e-04(10) \\ \hline 
car&8.90e-05(7) &1.00e-04(10) &5.86e-04(15) &1.03e-04(11) &3.65e-05(1) &9.79e-05(9) &5.24e-03(16) &1.39e-04(12) \\ \hline 
census-income&4.52e-04(9) &4.07e-04(4) &5.30e-04(13) &4.09e-04(6) &4.07e-04(5) &4.47e-04(8) &4.98e-04(10) &1.08e-01(17) \\ \hline 
cmc&8.76e-04(11) &1.89e-04(7) &5.84e-04(9) &5.81e-05(1) &9.60e-04(14) &6.01e-05(2) &2.33e-03(16) &9.2e-04(13) \\ \hline 
dota2&5.14e-04(10) &4.45e-04(6) &7.14e-04(13) &4.26e-04(2) &4.56e-04(7) &4.37e-04(4) &5.82e-04(12) &3.82e-02(16) \\ \hline 
drug-consumption&1.77e-04(1) &2.34e-04(2) &1.83e-03(15) &2.42e-04(3) &2.61e-04(5) &2.95e-04(8) &2.86e-03(16) &3.36e-04(11) \\ \hline 
fertility&8.51e-04(13) &1.93e-04(5) &1.00e-03(14) &5.13e-04(7) &5.28e-04(10) &5.15e-04(9) &2.94e-03(16) &6.23e-04(11) \\ \hline 
IndianLiverPatient&2.26e-04(10) &5.91e-05(2) &3.7e-04(13) &8.41e-05(3) &1.24e-04(6) &8.68e-05(4) &1.24e-03(16) &2.4e-04(11) \\ \hline 
iris&8.33e-04(11) &1.45e-04(2) &1.86e-03(12) &1.51e-04(3) &8.22e-04(10) &1.92e-03(15) &1.89e-03(14) &5.40e-04(9) \\ \hline 
krkopt&6.59e-05(5) &6.95e-05(9) &2.48e-03(16) &6.83e-05(8) &5.09e-05(1) &6.19e-05(3) &2.34e-03(15) &1.63e-03(14) \\ \hline 
letter-recognition&5.61e-05(2) &7.25e-05(5) &4.15e-03(16) &7.20e-05(4) &8.98e-05(7) &9.04e-05(8) &2.48e-03(15) &1.06e-03(13) \\ \hline 
magic04&8.97e-05(11) &6.80e-05(4) &2.21e-04(15) &6.76e-05(3) &4.79e-05(1) &5.07e-05(2) &1.47e-04(12) &4.46e-03(16) \\ \hline 
ml-prove&3.34e-04(10) &2.00e-04(1) &5.86e-04(16) &2.38e-04(3) &2.71e-04(7) &3.03e-04(9) &4.43e-04(12) &5.45e-04(15) \\ \hline 
musk1&1.21e-03(10) &1.07e-03(6) &2.56e-03(15) &9.35e-04(3) &8.49e-04(1) &1.05e-03(5) &2.78e-03(16) &1.21e-03(9) \\ \hline 
musk2&8.82e-04(6) &9.37e-04(10) &1.83e-03(14) &8.97e-04(7) &9.07e-04(9) &9.05e-04(8) &4.20e-03(16) &2.92e-03(15) \\ \hline 
optdigits&2.37e-04(2) &3.58e-04(5) &4.13e-03(16) &3.78e-04(8) &2.48e-04(3) &3.68e-04(7) &6.09e-04(13) &1.33e-03(15) \\ \hline 
page-blocks&9.29e-05(10) &5.12e-05(3) &7.04e-04(15) &5.32e-05(4) &4.16e-05(1) &7.07e-05(7) &6.82e-04(14) &1.92e-03(16) \\ \hline 
parkinsons&3.97e-04(8) &4.14e-04(10) &9.03e-04(15) &4.11e-04(9) &3.13e-04(5) &5.81e-04(12) &8.42e-04(14) &3.94e-04(7) \\ \hline 
pendigits&8.28e-05(1) &1.14e-04(9) &1.17e-03(15) &1.11e-04(8) &1.01e-04(7) &9.39e-05(2) &2.92e-04(12) &1.22e-03(16) \\ \hline 
segmentation&7.14e-04(2) &1.25e-03(15) &4.14e-03(16) &7.36e-04(3) &8.34e-04(5) &6.83e-04(1) &9.66e-04(7) &1.2e-03(11) \\ \hline 
sensorless&2.91e-04(8) &2.22e-04(5) &4.21e-03(14) &1.83e-04(3) &2.15e-04(4) &1.35e-04(1) &1.09e-03(12) &1.22e-02(16) \\ \hline 
tae&8.21e-04(13) &1.51e-04(3) &2.38e-03(16) &1.48e-04(2) &2.75e-04(5) &1.44e-04(1) &1.74e-03(15) &2.81e-04(7) \\ \hline 
wine&2.88e-04(6) &1.25e-04(3) &2.30e-03(16) &1.22e-04(2) &3.88e-04(9) &5.04e-04(11) &5.90e-04(12) &2.83e-04(5) \\ \hline 
yeast&2.31e-04(8) &1.2e-04(2) &1.31e-03(16) &2.44e-04(10) &1.54e-04(3) &1.15e-04(1) &5.97e-04(14) &4.02e-04(13) \\ \hline 
Average time per sample &0.0087&0.0089&0.0091&0.0097&0.0099&0.0099&0.0105&0.0172\\ \hline 
Average rank &7.9&5.9&13.8&5.4&5.9&6.6&13.9&13.0\\ \hline 
\end{tabular}
\label{tab:res_apx_classifiers_time_part2}
\end{table}
\end{landscape}

\begin{landscape}
\begin{table}[htbp!]
\supertiny
\centering
\caption[Comparison of EvoDAG against state-of-the-art classifiers based on time, the ranks are in parenthesis.]{Performance using time (with ranks) of: LogisticRegression (LR), LinearSVC (LSVC), SVC, SGDClassifier (SDG), PassiveAggressiveClassifier (PA), DecisionTreeClassifier (DT), ExtraTreesClassifier (ET), RandomForestClassifier (RF), AdaBoostClassifier (AB) and GradientBoostingClassifier (GB). Part 3.}
\begin{tabular}{| l |r|r|r|r|r|r|r|r|r|r|r|r|r|r|}
\hline
 & EvoDAG rnd-rnd & EvoDAG fit-fit & EvoDAG agr-rnd & EvoDAG nvs-rnd & EvoDAG ads-rnd & autosklearn & tpot \\ \hline \hline
ad&3.25e-01(17) &5.17e-01(18) &8.48e-01(19) &4.91e+00(22) &2.4e+00(21) &1.57e+00(20) &2.82e+02(23) \\ \hline 
adult&2.16e-01(19) &3.47e-01(20) &8.44e-01(21) &2.05e-01(18) &2.40e+00(22) &1.10e-01(17) &9.85e+00(23) \\ \hline 
agaricus-lepiota&4.56e-01(17) &5.27e-01(18) &8.36e-01(20) &3.53e+00(21) &6.17e+00(22) &6.32e-01(19) &1.14e+01(23) \\ \hline 
aps-failure&1.39e-01(2) &1.23e-01(1) &5.71e-01(21) &3.70e-01(20) &1.99e+00(22) &1.58e-01(3) &3.38e+01(23) \\ \hline 
banknote&2.82e-02(17) &1.34e-01(19) &8.38e-02(18) &4.25e-01(20) &1.81e+00(21) &3.75e+00(22) &4.43e+00(23) \\ \hline 
bank&1.4e-01(18) &1.95e-01(20) &5.21e-01(21) &1.80e-01(19) &1.91e+00(22) &1.14e-01(17) &1.40e+01(23) \\ \hline 
biodeg&4.36e-01(18) &3.86e-01(17) &8.12e-01(19) &2.42e+00(20) &3.02e+00(21) &4.87e+00(22) &2.42e+01(23) \\ \hline 
car&3.18e-01(17) &5.5e-01(18) &8.44e-01(19) &2.85e+00(20) &4.83e+00(22) &2.97e+00(21) &3.07e+01(23) \\ \hline 
census-income&1.66e-01(19) &1.45e-01(18) &5.09e-01(21) &4.78e-01(20) &1.79e+00(22) &1.85e-02(16) &1.15e+01(23) \\ \hline 
cmc&3.28e-01(17) &5.04e-01(18) &6.93e-01(19) &2.75e+00(20) &3.29e+00(21) &3.49e+00(22) &2.37e+01(23) \\ \hline 
dota2&3.29e-01(19) &2.59e-01(18) &6.62e-01(21) &5.35e-01(20) &1.98e+00(22) &3.95e-02(17) &1.14e+01(23) \\ \hline 
drug-consumption&2.38e-01(17) &2.65e-01(18) &4.31e-01(19) &2.25e+00(20) &4.76e+00(22) &2.73e+00(21) &5.53e+01(23) \\ \hline 
fertility&1.24e+00(18) &1.65e+00(19) &1.20e+00(17) &3.01e+00(20) &3.94e+00(21) &5.21e+01(22) &7.42e+01(23) \\ \hline 
IndianLiverPatient&3.07e-01(17) &5.61e-01(18) &1.06e+00(19) &3.17e+00(20) &3.41e+00(21) &8.83e+00(22) &1.87e+01(23) \\ \hline 
iris&4.76e-01(17) &7.5e-01(18) &9.98e-01(19) &1.69e+00(20) &4.77e+00(21) &3.43e+01(22) &5.08e+01(23) \\ \hline 
krkopt&1.28e+00(18) &1.05e+00(17) &1.39e+00(19) &3.39e+00(20) &1.08e+01(21) &5.21e+01(22) &7.76e+01(23) \\ \hline 
letter-recognition&1.46e+00(18) &1.19e+00(17) &1.75e+00(19) &3.07e+00(20) &1.18e+01(21) &5.21e+01(23) &3.92e+01(22) \\ \hline 
magic04&3.95e-01(18) &4.19e-01(19) &1.30e+00(20) &4.38e+00(22) &2.91e+00(21) &2.70e-01(17) &5.18e+01(23) \\ \hline 
ml-prove&5.73e-02(17) &1.83e-01(19) &1.38e-01(18) &4.78e-01(20) &8.52e-01(22) &7.84e-01(21) &1.69e+01(23) \\ \hline 
musk1&6.55e-01(18) &6.22e-01(17) &9.56e-01(19) &3.10e+00(20) &3.48e+00(21) &1.08e+01(22) &2.82e+02(23) \\ \hline 
musk2&2.77e-01(17) &3.08e-01(18) &7.44e-01(19) &3.75e+00(22) &2.57e+00(21) &7.8e-01(20) &5.68e+01(23) \\ \hline 
optdigits&7.58e-01(17) &1.54e+00(20) &9.40e-01(18) &3.27e+00(21) &6.66e+00(22) &9.41e-01(19) &6.32e+01(23) \\ \hline 
page-blocks&1.46e-01(17) &1.77e-01(18) &3.3e-01(19) &1.86e+00(21) &3.01e+00(22) &9.39e-01(20) &3.98e+01(23) \\ \hline 
parkinsons&5.73e-01(17) &6.22e-01(18) &8.49e-01(19) &2.55e+00(20) &4.75e+00(21) &2.66e+01(22) &3.37e+01(23) \\ \hline 
pendigits&6.80e-01(18) &1.84e+00(20) &1.24e+00(19) &2.95e+00(21) &1.22e+01(22) &4.8e-01(17) &5.68e+01(23) \\ \hline 
segmentation&8.57e-01(17) &1.23e+00(19) &1.09e+00(18) &2.95e+00(20) &6.92e+00(21) &1.71e+01(22) &1.03e+02(23) \\ \hline 
sensorless&9.22e-01(18) &1.48e+00(19) &2.16e+00(20) &2.26e+00(21) &1.55e+01(22) &8.83e-02(17) &2.23e+01(23) \\ \hline 
tae&1.03e+00(17) &1.64e+00(18) &1.64e+00(19) &3.65e+00(20) &9.55e+00(21) &3.43e+01(22) &8.20e+01(23) \\ \hline 
wine&6.02e-01(17) &7.02e-01(18) &1.04e+00(19) &1.82e+00(20) &3.87e+00(21) &2.93e+01(22) &8.69e+01(23) \\ \hline 
yeast&4.68e-01(17) &5.61e-01(18) &8.27e-01(19) &3.02e+00(20) &8.11e+00(22) &3.46e+00(21) &6.25e+01(23) \\ \hline 
Average time per sample &0.5102&0.683&0.9104&2.3754&5.0449&11.5265&57.6855\\ \hline 
Average rank &17.0&17.8&19.2&20.3&21.5&19.7&23.0\\ \hline 
\end{tabular}
\label{tab:res_apx_classifiers_time_part3}
\end{table}
\end{landscape}
\small

\bibliography{bibliography}

\end{document}